\documentclass[lettersize,journal]{IEEEtran}
\usepackage{amsmath,amsfonts}
\usepackage{algorithmic}
\usepackage{algorithm}
\usepackage{array}
\usepackage[caption=false,font=normalsize,labelfont=sf,textfont=sf]{subfig}
\usepackage{textcomp}
\usepackage{stfloats}
\usepackage{url}
\usepackage{verbatim}
\usepackage{graphicx}
\usepackage{cite}
\usepackage[accsupp]{axessibility}
\usepackage{soul}

\usepackage{times}
\usepackage{epsfig}

\usepackage{amsmath}
\usepackage{amssymb}
\usepackage{booktabs}
\usepackage[table]{xcolor} %POL
\usepackage{tikz}
\usepackage{comment}
\usepackage{color}
\usepackage{multirow} % Pol
\usepackage{siunitx}  %
\usepackage{subfig}
\usepackage{float}

\usepackage[symbol]{footmisc}
\usepackage{pifont}

\usepackage{makecell}

\usepackage{multirow}
\usepackage{hyperref}

\newcommand{\pol}[1]{{\color{red}#1}}

\newcommand{\method}{SIRA++}
\newcommand{\priorgeo}{SA-SM}

\newcommand{\fsdf}{f^{\rm sdf}}

\newcommand{\bc}{\mathbf{c}}

\newcommand{\bn}{\mathbf{n}}

\newcommand{\br}{\mathbf{r}}

\newcommand{\bt}{\mathbf{t}}

\newcommand{\bv}{\mathbf{v}}

\newcommand{\bx}{\mathbf{x}}

\newcommand{\bz}{\mathbf{z}}

\newcommand{\bI}{\mathbf{I}}

\newcommand{\bM}{\mathbf{M}}

\newcommand{\bT}{\mathbf{T}}

\newcommand{\mC}{\mathcal{C}}

\newcommand{\mL}{\mathcal{L}}

\newcommand{\mP}{\mathcal{P}}

\newcommand{\mbR}{\mathbb{R}} %symbol for the Real numbers
\newcommand{\bdelta}{\boldsymbol{\delta}}

\newcommand{\bgamma}{\boldsymbol{\gamma}}

\newcommand{\btheta}{\boldsymbol{\theta}}

\newcommand{\argmin}{\operatornamewithlimits{arg\,min}}

\begin{document}

%\title{SIRA: Relightable Avatars From a Single Image}
\title{Implicit Shape and Appearance Priors for Few-Shot Full Head Reconstruction}

%Eduard Ramon\textsuperscript{2*}

%\author{Pol Caselles\textsuperscript{1,2}, Eduard Ramon$^{2 \textbf{*}}$\thanks{$^*$This work was done prior to joining Amazon.}, Jaime García\textsuperscript{2}, Gil %Triginer\textsuperscript{2}, Francesc Moreno-Noguer\textsuperscript{1} \\[0.5cm]
%\textsuperscript{1} Institut de Robòtica i Informàtica Industrial, CSIC-UPC, Barcelona, Spain\\
%\textsuperscript{2} Crisalix SA %\hspace{1cm} \textsuperscript{3} Amazon
%}

%\author{Pol Caselles\textsuperscript{1,2}, Eduard Ramon$^{2,3 \textbf{*}}$\thanks{$^{*}$ Work done prior  joining Amazon.}, Jaime García\textsuperscript{2}, Gil Triginer$^{2 \textbf{$\dagger$}}$\thanks{$^{\dagger}$Work   done prior  joining Astrazeneca.}, Francesc Moreno-Noguer\textsuperscript{1,3\textbf{*}}$ \\[0.5cm]
%\textsuperscript{1} Institut de Robòtica i Informàtica Industrial, CSIC-UPC, Barcelona, Spain\\
%\textsuperscript{2} Crisalix SA \hspace{1cm} \textsuperscript{3} Amazon
%}

\author{Pol Caselles\textsuperscript{1,2}, Eduard Ramon$^{2,3 \textbf{*}}$\thanks{$^{*}$ Work done prior  joining Amazon.}, Jaime García\textsuperscript{2}, Gil Triginer$^{2 \textbf{$\dagger$}}$\thanks{$^{\dagger}$Work   done prior  joining Astrazeneca.}, Francesc Moreno-Noguer\textsuperscript{1,3\textbf{*}} \\[0.5cm]
\textsuperscript{1} Institut de Robòtica i Informàtica Industrial, CSIC-UPC, Barcelona, Spain\\
\textsuperscript{2} Crisalix SA \hspace{1cm} \textsuperscript{3} Amazon
}

%\author{Pol Caselles,~@crisalix.com, Gil Triginer,~\IEEEmembership{Staff,~IEEE,}, Francesc Moreno Noguer,~\IEEEmembership{Staff,~IEEE,}
        % <-this % stops a space
%\thanks{This paper was produced by the IEEE Publication Technology Group. They are in Piscataway, NJ.}% <-this % stops a space
%\thanks{Manuscript received April 19, 2021; revised August 16, 2021.}}



\maketitle

%\vspace{-30mm}
\begin{abstract}
Recent advancements in learning techniques that employ coordinate-based neural representations have yielded remarkable results in multi-view 3D reconstruction tasks.  However, these approaches often require a substantial number of input views (typically several tens) and computationally intensive optimization procedures to achieve their effectiveness. In this paper, we address these limitations specifically for the problem of few-shot full 3D head reconstruction.  We accomplish this by incorporating a probabilistic shape and appearance prior into coordinate-based representations, enabling faster convergence and improved generalization when working with only a few input images (even as low as a single image). During testing, we leverage this prior to guiding the fitting process of a signed distance function using a differentiable renderer. By incorporating the statistical prior alongside parallelizable ray tracing and dynamic caching strategies, we achieve an efficient and accurate approach to few-shot full 3D head reconstruction.

Moreover, we extend the H3DS dataset, which now comprises 60 high-resolution 3D full-head scans and their corresponding posed images and masks, which we use for evaluation purposes. By leveraging this dataset, we demonstrate the remarkable capabilities of our approach in achieving state-of-the-art results in geometry reconstruction while being an order of magnitude faster than previous approaches. 
\end{abstract}

\begin{IEEEkeywords}
Neural Radiance Field, Signed Distance Function, Few-shot 3D Reconstruction
\end{IEEEkeywords}

\section{Introduction}

In recent years, the digitalization of humans has emerged as an important area of research. The ability to create accurate digital representations of individuals holds immense value across a wide range of applications, including Virtual Reality (VR), Augmented Reality (AR), healthcare, entertainment, and security.  
To achieve this, the process often entails capturing photographs of a scene using conventional cameras or mobile devices. However, generating 3D reconstructions under non-controlled conditions or with a limited number of input images can be particularly challenging~\cite{wu2019mvf,ramon2019multi,bai2020deep,dou2018multi}.  Among these challenges, the single-view setup stands out as the most ill-posed scenario. It involves a highly under-constrained problem that cannot be effectively solved without prior knowledge or additional information~\cite{tewari2017mofa,tuan2017regressing,richardson20163d,richardson2017learning,sela2017unrestricted,tran2018extreme}.

\vspace{4mm}
Statistical priors based on 3D Morphable Models \cite{bai2020deep, dou2018multi, Moreno_pami2013, ramon2019multi, richardson20163d, richardson2017learning, tewari2017mofa, tran2018extreme, tuan2017regressing, wu2019mvf} have become the standard approach for few-shot 3D face reconstruction. By adopting 3DMMs as a representation, the problem of 3D reconstruction can be simplified to estimating a small set of parameters that effectively capture a target 3D shape. This enables the generation of 3D reconstructions from small sets of images \cite{bai2020deep, dou2018multi, ramon2019multi, wu2019mvf},  even when only a single  view is available \cite{richardson20163d, richardson2017learning, tewari2017mofa, tran2018extreme, tuan2017regressing}. However, one significant drawback of morphable models is their limited expressiveness, particularly for capturing high-frequency details. To address this issue, researchers have explored post-processing techniques that transfer fine details from the image domain to enhance the 3D geometry \cite{Lin_2020_CVPR, richardson2017learning, tran2018extreme}. Another limitation of 3DMMs is their inability to represent complex shapes and diverse topologies. Consequently, they are not well-suited for reconstructing complete heads with features such as hair, beard, facial accessories, and upper body clothing.

Model-free approaches based on discrete representations such as voxels\cite{jackson2017large}, meshes or point clouds offer greater flexibility in representing a wide range of shapes. However, they come with computational limitations, as they face scalability challenges as resolution increases or are limited to fixed topologies. Despite significant advancements in computational resources in recent years, there remains a trade-off between resolution and memory. Overcoming these challenges, neural fields \cite{chen2019learning, mescheder2019occupancy, mildenhall2020nerf, park2019deepsdf,  pumarola2020d, saito2019pifu, sitzmann2019scene, yu2021pixelnerf} have emerged as a solution that encodes both geometry and appearance as a continuous coordinate-based function within the weights of a neural network. Notably, recent work by \cite{niemeyer2020differentiable,yariv2020multiview} has demonstrated the success of such representations in learning detailed 3D geometry directly from images, even in the absence of 3D ground truth supervision. Unfortunately, these methods currently rely on a significant number of input views, leading to time-consuming inference and limiting their applications.

Optimization-based techniques iteratively refine model parameters to minimize a cost function, resulting in high accuracy and fine detail capture \cite{martin2021nerf,neus2}. On the contrary, feed-forward methods highly rely on the quality and variability of the training data which limits their ability to reconstruct fine details in out-of-distribution samples at test time \cite{Morales2021-at}. Nonetheless, they offer the advantage of speed and computational efficiency, making them suitable for real-time applications \cite{MICA:ECCV2022,feng2021learning,zhu2023facescape, wang2022faceverse, hane2017hierarchical}.

\begin{figure*}[t!]
    \centering
    \includegraphics[width=1.\textwidth]{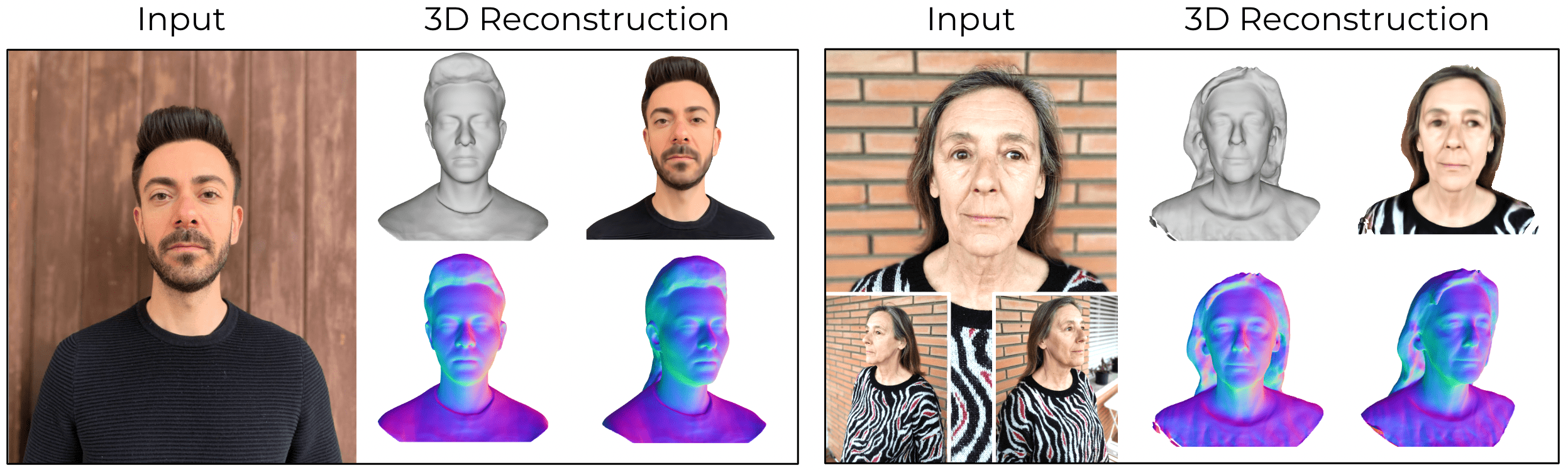}
    \vspace{-6mm}
    \caption{\textbf{Few-shot full-head reconstruction using \method{}}. Our approach enables high-fidelity 3D head reconstruction using only a few images. The figure showcases two examples, one obtained from a single input image (in 92 seconds) and the other from three input images (in 191 seconds). For each example, we present the input image/s on the left and the corresponding reconstruction on the right, including the 3D mesh, rendered mesh, and normal maps. The results demonstrate the effectiveness and efficiency of our method in generating detailed 3D head avatars with minimal input data. 
    }
    \label{fig:teaser}
    \vspace{-4mm}
\end{figure*}

Recent advancements in this field, highlighted by H3D-Net~\cite{ramon2021h3d} leverage large 3D scan datasets to integrate prior geometric knowledge into neural field models, allowing for significantly improved accuracy of full-head 3D reconstructions, even when only a limited number of input images are available. Nonetheless, it is worth noting that~\cite{ramon2021h3d} still faces challenges in terms of computational cost, demanding several hours of optimization per scene at inference time, and it is not suitable for scenarios where only a single input image is available.

To address the limitations of H3D-Net, we propose a novel prior that combines shape and appearance information. Our approach leverages an extensive dataset of 10.000 textured 3D head scans to pretrain an architecture consisting of two primary neural field decoders. The first decoder focused on modeling the complete geometry of the head using a signed distance function (SDF) as a 3D representation. The second decoder aims to capture the head appearance, including the face region, hair, and clothing of the upper torso. During inference, we utilize this pre-trained appearance and shape prior to initialize and guide the optimization of an Implicit Differentiable Renderer (IDR) \cite{yariv2020multiview} that, given a reduced number of input images, estimates the full head geometry. The learned prior facilitates faster convergence during optimization ($\sim$191 seconds per 3 input images) and prevents the model from getting trapped in local minima. As a result, our method produces 3D shape estimates that capture fine details of the face, head, and hair from just a single input image.

The core of our approach, dubbed SIRA, has been previously presented in~\cite{Caselles_2023_SIRA}. In this journal submission, we expand upon~\cite{Caselles_2023_SIRA} through innovative strategies to efficiently eliminate non-intersecting rays. These strategies, combined with a parallelizable ray tracing algorithm and dynamic caching, result in a remarkable acceleration of over $10\times$ compared to the previous implementation. We refer to this enhanced version of the approach as SIRA++.

Moreover, in this submission, we provide an in-depth analysis of SIRA++ for multi-view setup, including the joint shape and appearance priors, camera noise robustness, extended comparison with new 8 SOTA methods, as well as extensive evaluation in-the-wild CelebA-HQ dataset. In addition, we have made significant enhancements to the H3DS dataset, which consists of high-resolution 3D full head scans, images, masks, and camera poses, originally introduced in~\cite{ramon2021h3d}. Our expansion has considerably increased the dataset size from 10 to 60 samples, resulting in a more extensive and diverse collection of data. This enhanced H3DS dataset serves as the foundation for a thorough evaluation of the performance of SIRA++. We compare our method with H3D-Net~\cite{ramon2021h3d}, SIRA~\cite{Caselles_2023_SIRA}, and other mesh and field based state-of-the-art approaches.

The experimental results demonstrate that SIRA++ surpasses H3D-Net by a substantial margin in terms of reconstruction error of the full head. Moreover, SIRA++ achieves comparable performance to SIRA but with significantly lower computational cost. Additionally, we compare SIRA++ with recent approaches that are based on parametric models, solely providing a 3D reconstruction of the face region rather than the entire head. SIRA++ consistently demonstrates improved results compared to these methods.

In summary, this work builds upon our earlier versions of H3D-Net~\cite{ramon2021h3d} and SIRA~\cite{Caselles_2023_SIRA}, which were pioneering approaches in utilizing implicit functions for the reconstruction of full 3D human heads from a limited number of images. In this submission, we enhance the reconstruction accuracy of H3D-Net by introducing a novel data-driven prior that combines shape and appearance. Additionally, we significantly improve the computational efficiency of SIRA by implementing parallelizable ray tracing and dynamic caching strategies. These advancements result in an algorithm that is highly efficient and accurate for full 3D head reconstruction, even when only a few images are available (including the most challenging case of a single image). Fig.~\ref{fig:teaser} shows two examples of the reconstructions we obtain from one  (in 92 seconds) or five  (in 191 seconds) input images.

Furthermore, this submission includes an extended version of the H3DS dataset, which we will make publicly available for evaluation purposes. This expanded dataset will provide researchers and practitioners with a valuable resource for assessing and benchmarking their algorithms and methodologies in the field of 3D head reconstruction.

\section{Related Work}

In this section, we review the related work on face and full head reconstruction,  with a primary focus on statistical morphable models and recent advancements in neural fields for 3D reconstruction.  Note that while there exists extensive literature on 3D body reconstructions, our paper specifically narrows its scope to those dealing with the full head region.

\begin{figure*}[!t]
    \centering
    \includegraphics[width=\textwidth]{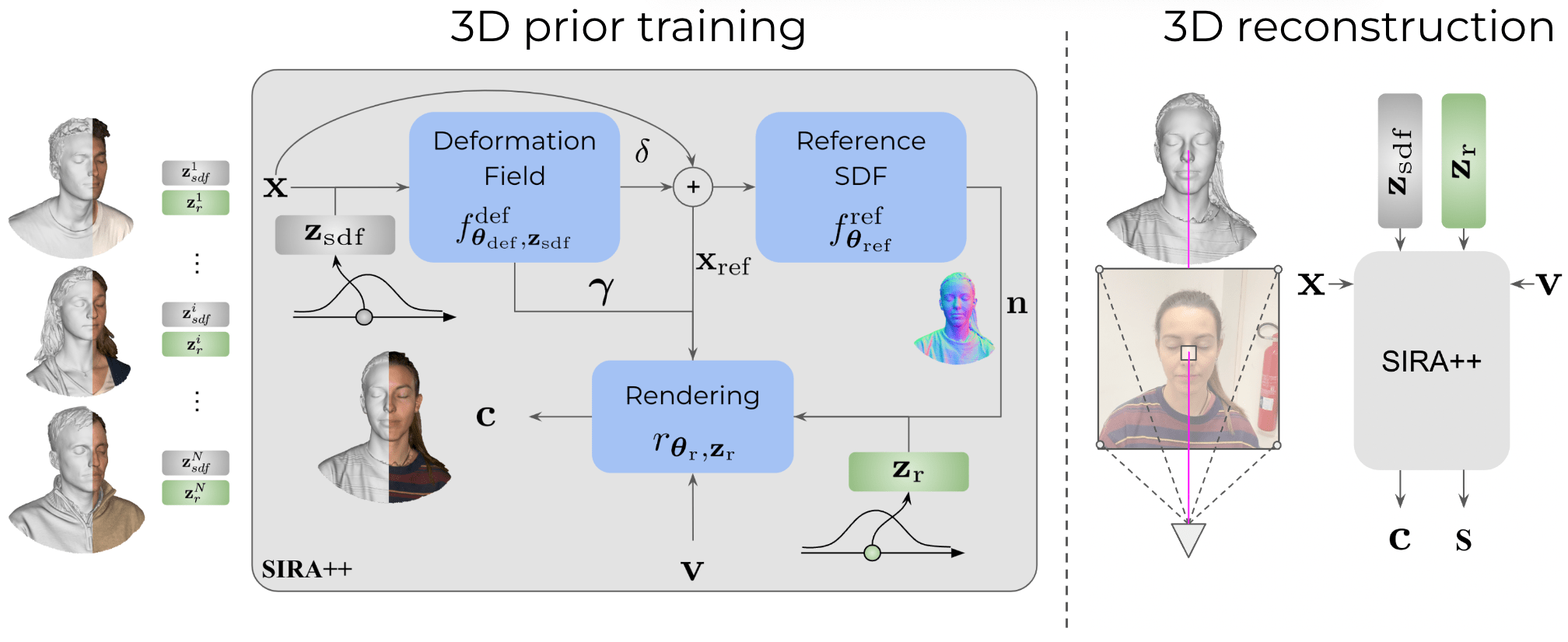}
    \vspace{-8mm}
    \caption{\textbf{Overview of \method{}.} \textbf{Left:} We construct a surface appearance statistical model using a dataset of raw head scans paired with multiview posed images. This involves learning a codebook of shapes $\bz_{\rm sdf}$ and appearances $\bz_{\rm rend}$, alongside two decoders that approximate a signed distance function and a renderer. The prior is trained using an autodecoder approach.
    \textbf{Right:} The pre-trained prior model is integrated with the implicit differentiable renderer. To begin the optimization process with a plausible human head, we sample from the manifold of shape and appearance latents. During the initial iterations, our focus is on training the latents to approximate the closest human head within our statistical model. Subsequently, we unfreeze the deformation and rendering networks, enabling fine-tuning of the fine details. Throughout the entire optimization phase, the reference network remains frozen, ensuring consistent results. }
    \label{fig:method}
    \vspace{-4mm}
\end{figure*}

\vspace{1mm}
\noindent\textbf{Statistical morphable models.}  The utilization of 3D Morphable Models (3DMMs) has become the predominant paradigm for face reconstruction from images, particularly in single-view or few-shot scenarios. These statistical models \cite{paysan20093d, FLAME:SiggraphAsia2017, HIFI3D} are widely adopted and primarily focus on the face region. In the context of single-shot methods, remarkable capabilities in face reconstruction have been demonstrated by \cite{tewari2017mofa, feng2021learning, MICA:ECCV2022, wang2022faceverse, zhu2023facescape}, even in challenging in-the-wild scenes. In an alternative approach, \cite{gecer2021fast, Gecer_2019_CVPR} employ a generative adversarial network for 3D face fitting. Recent works have also employed 3DMMs to estimate 3D geometry and spatially varying surface properties, such as diffuse and specular albedos, along with global illumination properties \cite{feng2021learning, dib2021towards, dib2021towards}. 

Achieving higher fidelity face reconstructions can be accomplished by leveraging multiple input images~\cite{wu2019mvf, bai2020deep, lei2023hierarchical}. Previous approaches have predominantly focused on optimizing model parameters through a multi-view analysis-by-synthesis strategy \cite{feng2021learning, dib2021towards}. To enhance robustness in specific face areas, \cite{li2023robust} utilizes occlusion cues. Furthermore, \cite{MICA:ECCV2022} proposes learning the statistical model within a metric space to effectively capture variations in scale.  Other works \cite{EMOCA:CVPR:2021} introduce emotional information to improve expression capture. Nevertheless, While these models have advanced 3D face reconstruction, they struggle to capture fine anatomical details and they are limited to the face region.

\vspace{1mm}
\noindent\textbf{Neural fields for 3D reconstruction.} 
In recent years, neural fields have emerged as the leading approach for scene representation~\cite{xie2021neural}, yielding remarkable results in novel view synthesis~\cite{mildenhall2020nerf, martin2021nerf, park2021nerfies, pumarola2021d, sitzmann2019scene} and 3D reconstruction~\cite{niemeyer2020differentiable, ramon2021h3d, yariv2020multiview, zhang2021physg}. These techniques have found practical applications in modeling full-head avatars~\cite{park2021nerfies, park2021hypernerf, zheng2021avatar, gafni2021dynamic}. By leveraging surface priors \cite{park2019deepsdf} and surface rendering \cite{yariv2020multiview}, neural fields enable highly accurate 3D reconstructions of the full head, encompassing intricate details such as hair and shoulders \cite{ramon2021h3d, zheng2021avatar}.\cite{lin2023single} employs a similar approach to construct implicit morphable faces with consistent texture parameterization and introduces single-shot inversion to obtain reconstructions from input images.  This method, however, requires substantial computational time, with a single scene taking approximately 3 hours to process.

In an effort  to enhance the geometric detail of morphable models, several approaches have combined them with implicit representations.  For example, \cite{zheng2022imface} enhances a morphable model representation through the use of an implicitly learned displacement field. However, this approach is limited to the face region. On the other hand, full head reconstruction is achieved in~\cite{grassal2022neural} through a feed-forward network that learns vertex displacements throughout the entire head. Another hybrid representation is proposed in \cite{zheng2022IMavatar}, which combines the fine-grained control mechanism of 3DMMs with the high-quality representation of implicit functions parametrized by neural networks. This approach enables the generation of animatable full-head avatars from videos. In an effort to expedite the training and rendering process, \cite{Zheng2023pointavatar} introduces a deformable point-based representation. Unfortunately, all of these approaches still rely on a significant number of input frames, which diverges from the few-shot or single-shot scenarios considered in this paper.

Recently, model-free approaches in combination with pixel-aligned features \cite{saito2019pifu,saito2020pifuhd,he2020geo,alldieck2022photorealistic,shao2022doublefield,corona2023structured} have emerged as an approach to obtain fast reconstructions as they don't require test-time optimization. PIFU \cite{saito2019pifu} introduced the concept of pixel-aligned features to condition a learned occupancy field from single to multiple input images and Phorum \cite{alldieck2022photorealistic} extended it by using a signed distance field as a shape representation. JIFF \cite{cao2022jiff} combines features from a face morphable model to enhance high-frequency details and KeypointNeRF \cite{Mihajlovic:ECCV2022} aggregates pixel-aligned features with a relative spatial encoder using volumetric rendering. However, pixel-aligned single feed-forward methods are still behind optimization-based approaches in terms of quality reconstruction. As we will show in the experimental section, these methods do not always guarantee realistic and accurate full-head reconstructions and are sensitive to camera pose estimation as they don't optimize cameras at test time.

\section{Method}
\label{sec:method}

\subsection{Problem Formulation}

Our objective is to recover the 3D head surface, denoted as $\mathcal{S}$, from a small set of $N\geq1$ input images $\bI_v$, where $v=1,\ldots,N$. Each input image is accompanied by its respective head mask $\bM_v$ and camera parameters $\bT_v$. We represent the surface $\mathcal{S}$ as the zero-level set of a signed distance function $\fsdf: \bx \rightarrow s$, such that $ \mathcal{S} = \{\bx \in \mbR^3 \mid \fsdf(\bx)=0\}$. After estimating $\fsdf$ from the visual cues $\bI_v$, $\bM_v$ and $\bT_v$, the surface $\mathcal{S}$ can be obtained as a post-processing step using Marching Cubes \cite{lorensen1998marching}

To tackle the inherent underconstrained nature of recovering 3D geometry from image data, the incorporation of regularizations is crucial in resolving existing ambiguities. The problem becomes increasingly challenging as the number of available images decreases. We specifically address this challenge within a range of 1 to 32 views, where the one-shot regime poses the greatest difficulty due to the absence of multi-view cues to disentangle geometric and color information. To tackle this challenge, we propose a novel architecture that capitalizes on both geometric and appearance priors. By harnessing these priors, our approach achieves precise 3D reconstructions even in scenarios lacking multi-view consistency.

We recover the 3D geometry from the input images through analysis-by-synthesis with differentiable surface rendering as in \cite{yariv2020multiview,ramon2021h3d}.  Our proposed architecture addresses the challenge of limited multi-view information by leveraging two key inductive biases. Firstly, we decompose the signed distance function $\fsdf$ into a reference SDF and a deformation field \cite{yenamandra2021i3dmm}. This parameterization serves as an implicit bias, ensuring that the composed SDF remains close to the reference. Secondly, we create a statistical prior that models the variations of shape and appearance of 3D head surfaces. This model is used during inference to achieve a reliable initialization and faster convergence. To enhance the robustness of the analysis-by-synthesis process during inference, we optimize the parameters of this statistical model, drawing inspiration from \cite{ramon2021h3d}, to achieve a reliable initialization. These inductive biases significantly enhance the performance of \method{}, surpassing that of \cite{ramon2021h3d}.

\subsection{Surface Appearance Statistical Model (SA-SM)} \label{sec:statistical_model}
In order to learn a statistical model capturing head shapes and appearances (Fig. \ref{fig:method}), we curate a dataset comprising scenes that contain raw head scans along with corresponding multiview posed images. For each scene, denoted by an index $i=1\dots M$, we extract a collection of surface points $\bx \in \mP_{\rm s}^{(i)}$ along with their respective normal vectors $\bn$. We project each surface point onto the images where it is visible, obtaining a set denoted as $\mC^{(i)}_\bx = {(\bc, \bv)}$, which consists of pairs comprising the associated RGB color $\bc$ and the corresponding viewing direction $\bv$.

\vspace{1mm}
\noindent{\textbf{\priorgeo{} Architecture:}} 
Our architecture consists of two primary neural field decoders: an SDF decoder denoted as  $f^{\rm sdf}_{\btheta_{\rm sdf}, \bz_{\rm sdf}}$, and a rendering function decoder represented as $r_{\btheta_{\rm r}, \bz_{\rm r}}$. Here, $\bz_{\rm sa} = \{\bz_{\rm sdf}, \bz_{\rm r}\}$ refers to the latent vectors encompassing the shape and appearance spaces of the \priorgeo{}. Additionally, $\btheta_{\rm sa} = \{\btheta_{\rm sdf}, \btheta_{\rm r}\}$ denotes the parameters associated with their respective decoders.

The SDF decoder consists of two sub-functions: a deformation field and a reference SDF. Our experimental results (Section \ref{ablation}, Figure \ref{fig:h3dnet2sira} and Table \ref{table:ablation_h3dnet2sira}) demonstrate that this separation introduces an implicit bias, effectively limiting significant deviations from the reference SDF. As a result, it enhances the stability of few-shot 3D reconstructions. The deformation field, is mathematically defined as:
\begin{equation}
    \label{eq:one_shot_def_net}
    f^{\rm def}_{\btheta_{\rm def}, \bz_{\rm sdf}}: \mathbb{R}^3 \rightarrow \mathbb{R}^{3+N_\gamma} \hspace{1mm},\hspace{1mm} \bx \mapsto (\bdelta, \bgamma),
\end{equation}
is parameterised by internal parameters $\btheta_{\rm def}$ and the latent vector $\bz_{\rm sdf}$. It maps input coordinates, $\bx$, to a deformation 3-vector, $\bdelta$.  Additionally, it generates an auxiliary feature vector $\bgamma$ of dimension $N_\gamma$, which encodes higher-level geometric information utilized by the differentiable renderer \cite{yariv2020multiview}. Note, however, that the rendering network does not include global lightning effects such us secondary lightning and self-shadows, as it is only conditioned on position, viewing direction, normals, and the appearance latent.

The predicted deformation is utilized to map an input coordinate $\bx$ to a coordinate $\bx_{\rm ref}$ within a reference space. In this reference space, we evaluate a reference signed distance function (SDF) $f^{\rm ref}_{\btheta_{\rm ref}}$, which is parameterized by internal parameters $\btheta_{\rm ref}$. This mapping process is expressed as follows:
\begin{eqnarray} 
\bx_{\rm ref} &=& \bx + \bdelta, \\
f^{\rm ref}_{\btheta_{\rm ref}}&:&\mathbb{R}^3\rightarrow \mathbb{R} \hspace{1mm} , \hspace{1mm} \bx_{\rm ref} \mapsto s. \label{one_shot_canonical_coord}
\end{eqnarray}

Combining the components described above, we obtain the composed SDF decoder:
\begin{equation}
    \label{eq:one_shot_geo_net}
    f^{\rm sdf}_{\btheta_{\rm sdf}, \bz_{\rm sdf}} : \bx \mapsto f^{\rm ref}_{\btheta_{\rm ref}}(\bx^{\rm ref}),
\end{equation} 
where the decoder internal parameters are $\btheta_{\rm sdf} = (\btheta_{\rm def}, \btheta_{\rm ref})$. The second main component of our architecture is the rendering function:
\begin{equation}
    \label{eq:one_shot_render_net}
    r_{\btheta_{\rm r}, \bz_{\rm r}}: (\bx_{\rm ref}, \bn, \bv, \bgamma) \mapsto \bc \;,
\end{equation}
which is parameterised by internal parameters $\btheta_{\rm r}$ and a latent vector $\bz_{\rm r}$. This function assigns an RGB color $\bc$, to each combination of 3D coordinates in the reference space $\bx_{\rm ref}$, unit normal vector $\bn$, and unit viewing direction vector $\bv$ in the real space.

 \begin{figure}[t!]
    \centering
    \includegraphics[width=\columnwidth]{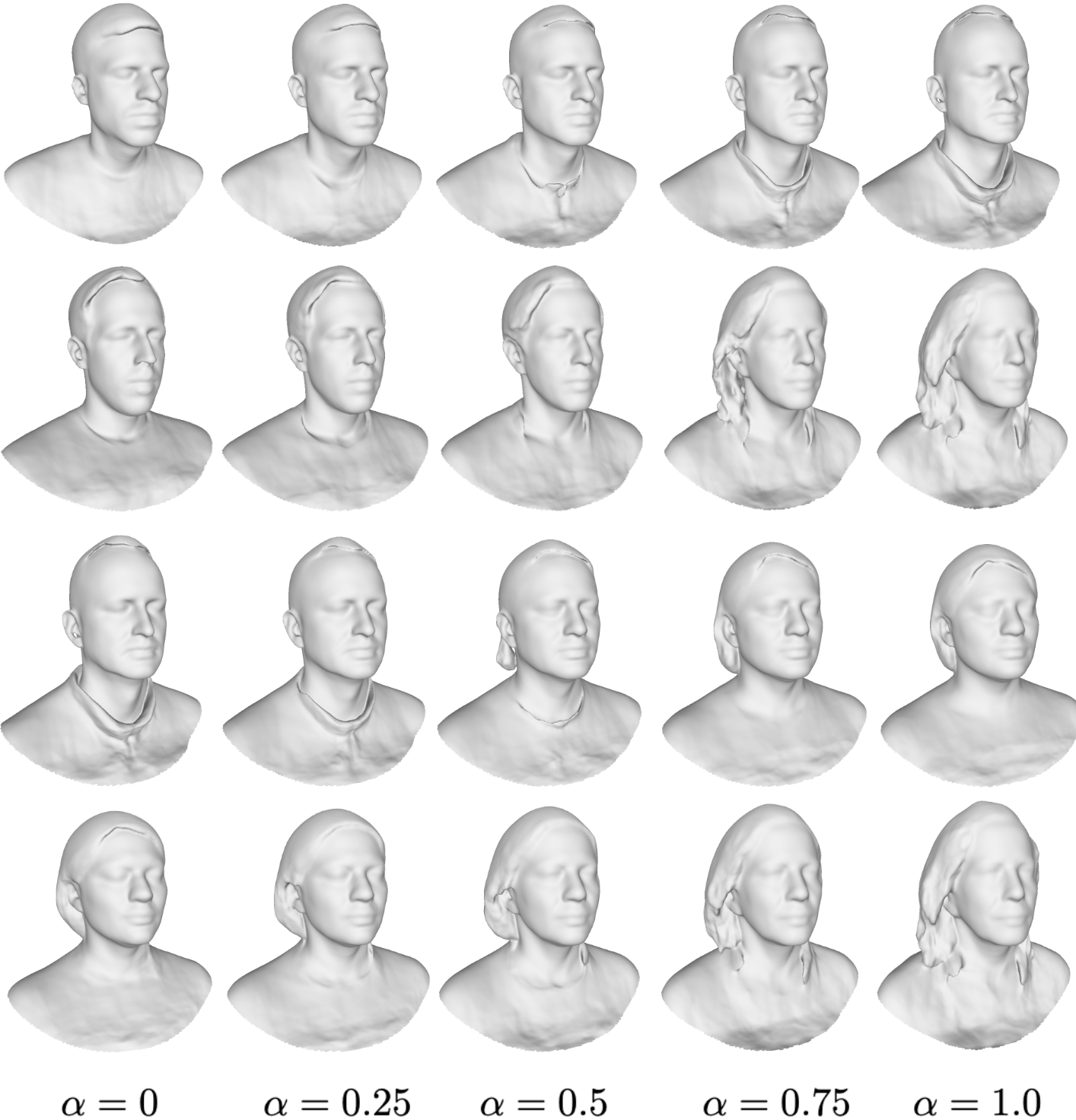}
    \vspace{-8mm}
    \caption{{\bfseries Latent shape interpolation.} 
    Each row of the figure depicts a latent interpolation between different subjects, controlled by a weight $\alpha$. This interpolation process showcases the smooth and gradual transformation of shapes, reflecting the continuous variation in human head representations along the latent space.
    }
    
    \label{fig:geo_interpolation}
    \vspace{-4mm}
\end{figure}

\vspace{1mm}
\noindent{\textbf{\priorgeo{} Training:}}
 To train our \priorgeo{} we adopt an auto-decoder framework in which each scene is associated with a set of latents $\bz_{\rm sa}^{(i)} = \{\bz_{\rm sdf}^{(i)}, \bz_{\rm r}^{(i)}\}$. These latents are optimized alongside the statistical model parameters $\btheta_{\rm sa}$. Upon completion of training, we obtain the optimized parameters $\btheta_{\rm sa,0} = \{\btheta_{\rm sdf,0}, \btheta_{\rm r,0}\}$ which establish that any combination of latents   $(\bz_{\rm sdf}, \bz_{\rm r})$ within the latent space corresponds  to a well-behaved SDF, $f^{\rm sdf}_{\btheta_{\rm sdf,0}, \bz_{\rm sdf}}$ and appearance $f^{\rm rend}_{\btheta_{\rm r,0}, \bz_{\rm r}}$ of a human head. To simplify notation and reduce complexity, we subsequently omit the dependency on the decoder's internal parameters.

In order to capture the space of head shapes, we sample a set of points on the surface of each training scan denoted as $\mP_{\rm s}^{(i)}$. Subsequently, we compute the surface error loss given by:  
\begin{equation}
    \mathcal{L}_{\rm Surf}^{(i)} = \sum_{\bx_j \in \mathcal{P}_{\rm s}^{(i)}} |f^{\rm sdf}_{\bz^{(i)}_{\rm sdf}}(\bx_j)|.
\end{equation}

In addition, we uniformly sample another set of points from the scene volume, $\mP_{\rm v}^{(i)}$, and compute the Eikonal loss \cite{gropp2020implicit}:
\begin{equation}
    \mathcal{L}_{\rm Eik}^{(i)} = \sum_{\bx_k \in \mathcal{P}_{\rm v}^{(i)} }  (\lVert\nabla_{\bx} f^{\rm sdf}_{\bz^{(i)}_{\rm sdf}}(\bx_k)\rVert-1 )^2.
\end{equation}

To encourage small-magnitude and zero-mean deformations, we incorporate a regularization term that prevents solutions where the deformations unnecessarily compensate for offset or scaled reference SDFs. The regularization term is defined as:
\begin{equation}
\begin{split}
\mL_{\rm Def}^{(i)} = \frac{1}{|\mP_{\rm s}^{(i)}|} \biggl( \sum_{\bx_j \in \mP_{\rm s}^{(i)}} \lVert \bdelta_{j}^{(i)}\rVert_2 + \biggl\lVert{\sum_{\bx_j \in \mP_{\rm s}^{(i)}} \bdelta_{ j}^{(i)}}\biggr\rVert_2\biggr),
\end{split}
\end{equation}
where $\bdelta_{j}^{(i)}$ represents the deformation vector applied to the 3D point $\bx_j$ within the scene indexed by  $i$.

Similar to the approach in~\cite{yenamandra2021i3dmm}, we employ a landmark consistency loss to ensure consistency among the 3D face landmarks. For each scene $i$, we automatically annotate a set of 3D face landmarks denoted as $\{\bx_l^{(i)}\}$ where $l=1 \dots L$. We then define the following loss that measures their deformed coordinate mismatch between pairs of scenes:
\begin{equation}
    \mL_{\rm Lm}^{(i)} = \sum_{j\neq i} \sum_{l}^{L} \lVert\bx_{\rm ref,l}^{(i)} - \bx_{\rm ref,l}^{(j)}\rVert^2\;,
\end{equation}
where $\bx_{\rm ref,l}^{(i)}$ represents the position of landmark $l$ of scene $i$ in the reference space.

The \priorgeo{} model learns a distribution of head appearances from the posed images associated with each training scene. To evaluate the rendering function (Eq.~\ref{eq:one_shot_render_net}), we calculate the coordinates in the reference space $\bx_{\rm ref}$ corresponding to the surface point (Eq. ~\ref{one_shot_canonical_coord}), along with the high-level descriptor $\bgamma$ (Eq.~\ref{eq:one_shot_def_net}). Additionally, we extract the surface normals, $\bn$, by normalizing the gradient of the SDF~\cite{yariv2020multiview}. Using these components, we define the color loss as follows:
\begin{equation}
    \label{eq:geometry_prior_color_loss}
    \mL_{\rm Col}^{(i)} = \sum_{\bx \in \mP_{\rm s}^{(i)}}\sum_{(\bc, \bv)\in\mC_\bx^{(i)}}  \lVert r_{\bz^{(i)}_{\rm r}}(\bx_{\rm ref}, \bn, \bv, \bgamma) - \bc \rVert \;.
\end{equation}

Finally, the $\mathcal{L}_{\rm Emb}^{(i)}$ term enforces a zero-mean multivariate-Gaussian distribution
with spherical covariance of $\sigma^2$ over the spaces of shape and appearance latent  vectors: 
$\mathcal{L}_{\rm Emb}^{(i)} = \frac{1}{\sigma^{2}}\bigl(\lVert\bz^{(i)}_{\rm sdf} \rVert_2 + \lVert\bz^{(i)}_{\rm r} \rVert_2\bigr).$
Combining all the loss terms, we minimize the following objective:
\begin{equation} \label{eq:deep_sdf_objective_color}
\begin{split}
\argmin_{\{\bz^{(i)}_{\rm sa}\},  \btheta_{\rm sa}} \sum_i \mL_{\rm Surf}^{(i)}  + \lambda_1\mL_{\rm Eik}^{(i)} + \lambda_2\mL_{\rm Def}^{(i)}+ \lambda_3\mL_{\rm Lm}^{(i)} + \\ \lambda_4\mL_{\rm Col}^{(i)}  +  \lambda_5\mL_{\rm Emb}^{(i)}
\end{split}
\end{equation}
where $\lambda_{1-5}$ are scalar hyperparameters.

\begin{figure}[t!]
    \centering
    \includegraphics[width=\columnwidth]{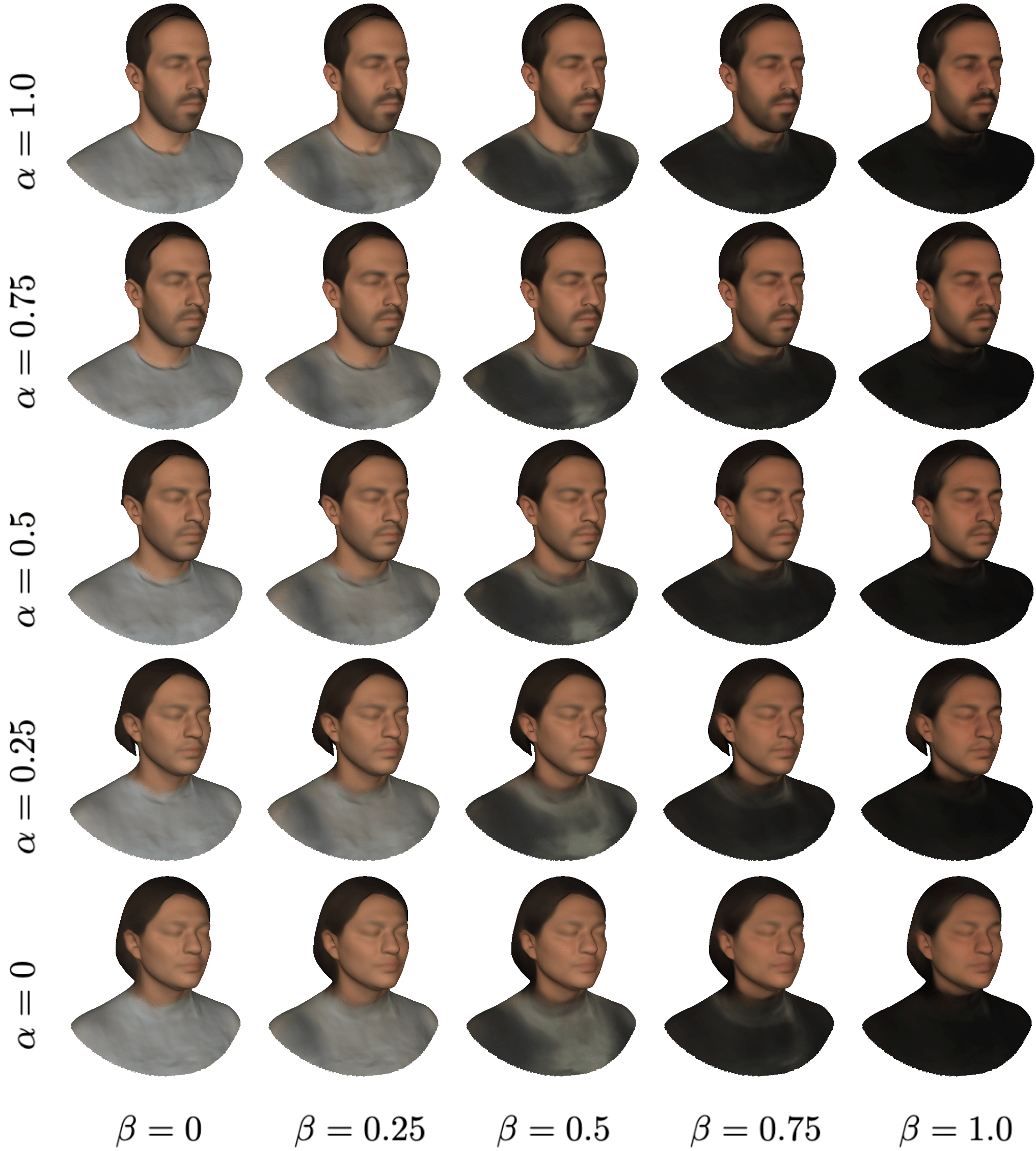}
    \vspace{-7mm}
    \caption{{\bfseries Latent shape and appearance interpolation.} This figure shows the joint shape and appearance latent interpolation between two subjects (top-left and bottom-right). Shape interpolation of the latent $\bz_{sdf}$ is controlled by means of the weight $\alpha$. Appearance interpolation of  $\bz_r$ is controlled by $\beta$.
    }
    \label{fig:color_interpolation}
    \vspace{-4mm}
\end{figure}

\vspace{1mm}
\noindent{\textbf{Expressivity of the SA-SM Prior:}} In order to assess the representation power of the shape and appearance prior we have learned, we conduct simple experiments by fitting our model to unseen subjects and interpolating their latent codes. In Fig.~\ref{fig:geo_interpolation}, we illustrate this process specifically for the shape prior. The ability to represent diverse fitting subjects indicates a rich and expressive manifold that captures diverse human head variations. Notably, the latent interpolation between different subjects (even across different genders) results in a remarkably smooth transition of feasible human heads. This observation highlights the structure and continuity of the learned latent codes. In Fig.~\ref{fig:color_interpolation}, we present a similar experiment where we simultaneously interpolate in both the shape space (vertical direction of the figure, governed by the weight $\alpha$) and the appearance space (horizontal direction, controlled by the weight $\beta$). Once again, we observe that our learned shape-and-appearance prior gracefully transitions between the two subjects represented at the top-left and bottom-right corners of the figure.  Interestingly, we observe that while the appearance latent effectively captures most of the color variance, such as the t-shirt, certain details like the mustache and beard are better represented by the geometry latent. This phenomenon arises because \method{} jointly models a distribution of 3D shapes and appearances, enabling the geometry latent to explain color variations that are statistically correlated with the underlying geometry.

\subsection{Geometry Reconstruction}

With our pre-trained statistical model at hand, we can tackle the task of obtaining a 3D reconstruction and an appearance from portrait images $\bI_v$ with associated camera parameters $\bT_v$ and foreground mask $\bM_v$.

To obtain precise 3D reconstructions of new scenes, our approach involves rendering the geometry described by $f^{\rm sdf}$ using the differentiable rendering function $r$ from Eq.~\ref{eq:one_shot_render_net}, while minimizing a photoconsistency error. Here's a step-by-step breakdown of our process:

First, for a given pixel coordinate $p$ in the input image $\bI_v$, we sample a ray $\br = {\bt + k\bv | k \geq 0}$, where $\bt$ represents the position of the associated camera $\bT_v$, and $\bv$ is the viewing direction. We then find the intersection coordinates of this ray with the composed SDF (Eq. \ref{eq:one_shot_geo_net}).

Next, we make this intersection point differentiable with respect to $\bz_{\rm sdf}$ and $\btheta_{\rm sdf}$ through implicit differentiation \cite{niemeyer2020differentiable, yariv2020multiview}. The resulting differentiable intersection coordinates $\bx_{\rm s}$ enable us to obtain their associated 3D displacement $\bdelta$ and feature vector $\bgamma$ (Eq. \ref{eq:one_shot_def_net}), as well as their corresponding coordinates in the reference space $\bx_{\rm ref}$ (Eq. \ref{one_shot_canonical_coord}), along with the normal vector $\bn = \nabla_\bx f^{\rm sdf}$.

\begin{table}[t]
\begin{center}
\caption{Testing Time  Comparison on H3DS Dataset with 1 and 3 Input Views. SS stands for Selective Sampling.}
\vspace{-4mm}
\label{tab:time_perf}
\begin{tabular}[t]{cccc}
\toprule
& Test time & Test time & Time \\
& 1 view $\downarrow$ & 3 views $\downarrow$ & reduction $\uparrow$\\
\cmidrule{1-4}

MVFNet \cite{wu2019mvf} & -& $<10s$ & - \\
DFNRMVS \cite{bai2020deep} & -& $<10s$ & - \\
DECA \cite{feng2021learning} & $<10s$ & - & - \\
MICA \cite{MICA:ECCV2022} & $<10s$ & - & - \\
FaceScape \cite{zhu2023facescape} & $<10s$ & - & - \\
FaceVerse \cite{wang2022faceverse} & 720s & - & - \\
HRN \cite{hane2017hierarchical} & $<5s$ & $ - $ & - \\
PIFU \cite{saito2019pifu} & $<10s$ & $<10s$ & - \\
JIFF \cite{cao2022jiff} & $<10s$ & $<10s$ & - \\
IDR \cite{yariv2020multiview} & -& $\sim1000s$ & - \\
NeuS2 \cite{neus2} & - & $\sim300s$ & - \\
H3D-Net \cite{ramon2021h3d} & - & $1050s$ & - \\
SIRA \cite{Caselles_2023_SIRA} & - & $3663s$ & - \\
 \cmidrule{1-4}
\method{} (w/o SS, w/o Cache) & - & $379s$ & 0\% \\
\method{} (w SS, w/o Cache) & - & $275s$ & 27\% \\
\method{} (w/o SS, w Cache) & - & $232s$ & 39\% \\
\method{} (w SS, w Cache) & - & \textbf{191s} & \textbf{50\%} \\
\bottomrule
\end{tabular}
\end{center}
\vspace{-4mm}
\end{table}

Finally, we compute the color associated with the ray as $\bc = r(\bx_{\rm ref}, \bn, \bv, \bgamma)$ using the differentiable rendering function $r$.

In order to optimize $\bz_{\rm sa}$ and $\btheta_{\rm sa}$, we minimize the following loss \cite{yariv2020multiview}:
\begin{equation} \label{eq:idr_loss}
\mathcal{L} = \mathcal{L}_{\rm RGB} + 
\lambda_{6}
\mathcal{L}_{\rm Mask} + 
\lambda_{7}\mathcal{L}_{\rm Eik},
\end{equation}
where $\lambda_{6}$ and $\lambda_{7}$ are hyperparameters.

We will now elaborate on each component of this loss. Let $\mathcal{P}$ be a mini-batch of pixels from the image $\bI_v$. We define $\mathcal{P}_{\rm RGB}$ as the subset of pixels whose associated ray intersects the surface defined by $f^{\rm sdf}$ and have a nonzero foreground mask value, while $\mathcal{P}_{\rm Mask} = \mathcal{P} \setminus \mathcal{P}_{\rm RGB}$. Let's delve into the specifics:

The first component, $\mathcal{L}_{\rm RGB}$, addresses photometric error, which is computed as follows:
\begin{equation}
    \mathcal{L}_{\rm RGB} = |\mP|^{-1} \sum_{p\in \mathcal{P}_{\rm RGB}} |\bI_v(p) - \bc_v(p)|.
\end{equation}

The second component, $\mathcal{L}_{\rm Mask}$, accounts for silhouette errors. It is defined as:
\begin{equation}
    \mathcal{L}_{\rm Mask} = \frac{1}{\lambda_{8}|\mathcal{P}|} \sum_{p \in \mathcal{P}_{\rm Mask}} {\rm CE}(\bM_v(p), s_{ \lambda_{8}}(p))\;,
\end{equation}
where $s(p) = {\rm sigmoid}(-\lambda_{8} \min_{t \geq 0} f^{\rm sdf}(\br_t))$ is the estimated silhouette. We use the binary cross-entropy $\rm CE$ to measure the difference between the foreground mask $\bM_v(p)$ and the estimated silhouette $s(p)$ for each pixel $p$ in $\mathcal{P}_{\rm Mask}$.

Lastly, $\mathcal{L}_{\rm Eik}$ encourages $f^{\rm sdf}$ to approximate a signed distance function.

Instead of optimizing all the parameters ${\btheta_{\rm sdf}, \btheta_{\rm r}, \bz_{\rm sdf}, \bz_{\rm r}}$ simultaneously, we propose a two-step schedule. First, we initialize the geometry and rendering functions with the parameters obtained from the pretraining described in the last section, denoted as ${\btheta_{\rm sdf,0}, \btheta_{\rm r,0}}$. The initial shape and appearance latents, $\bz_{\rm sdf}$ and $\bz_{\rm r}$, are sampled from a multivariate normal distribution with zero mean and a small variance, ensuring that they start near the mean of the latent spaces. In the first optimization phase, we exclusively optimize the shape and appearance latents. This yields an initial approximation within the previously learned shape and appearance latent spaces. Subsequently, in the second phase, we unfreeze the parameters of the deformation and rendering networks, denoted as ${\btheta_{\rm def}, \btheta_{\rm r}}$ (Eqs. \ref{eq:one_shot_def_net} and \ref{eq:one_shot_render_net}), while keeping the parameters of the reference Signed Distance Function (SDF), $\btheta_{\rm ref}$, frozen.

This two-step scheduling plays a pivotal role in achieving accurate results, particularly in the one-shot regime. By unfreezing the deformation and rendering networks, we can attain highly detailed solutions that lie outside of the pre-learned latent spaces. However, a crucial aspect of our approach is expressing the shape as a deformed reference Signed Distance Function (SDF), which acts as a regularization mechanism, ensuring proper training convergence.

The fine-tuned shape parameters resulting from this two-step process are denoted as $\btheta_{\rm def,ft}$ and $\bz_{\rm sdf, ft}$.

\section{Accelerating Reconstruction Geometry}

Reconstruction methods relying on surface rendering often require considerable time to find the intersection between cast rays and the reconstructed surface, leading to run times ranging from 15 minutes to several hours per scene~\cite{ramon2021h3d, yariv2020multiview, Caselles_2023_SIRA}. To address this challenge, we have introduced several enhancements in \method{}, including code optimization, adoption of a more parallelizable ray tracing algorithm~\cite{canela2023instantavatar}, implementation of a new scheduler to discard non-intersecting rays, and dynamic caching of the Signed Distance Function (SDF) during training. These improvements, as depicted in Table~\ref{tab:time_perf}, yield a substantial reduction in overall computation time, making our proposed method significantly faster and more efficient in comparison to existing techniques. In the following sections, we provide a detailed explanation of each of these enhancements.

\vspace{1mm}
\noindent\textbf{Code optimitzation.} 
SIRA++ has been developed within the PyTorch framework, with accelerated training on a single GPU. To optimize performance, we employ mixed precision during the ray tracing step, group query points into a single batch to minimize GPU calls, and minimize the use of memory copy instructions (e.g., reshape, concatenations, clones). Additionally, we strategically fuse operations, such as expressing sinus and cosines as a single call in the positional encoder. These measures collectively contribute to enhancing the efficiency and speed of the SIRA++ framework during the training process.

\vspace{1mm}
\noindent\textbf{Selective sampling.} 
In traditional 3D reconstruction methods based on surface or volumetric rendering, rays are often uniformly sampled on the image, leading to many rays inefficiently sampling empty space far from the surface. To enhance the efficiency of our reconstruction process, we implement a progressive strategy where we gradually stop sampling rays from the background during the optimization. This approach significantly reduces computational time while preserving the same level of reconstruction accuracy. By intelligently focusing on rays closer to the surface of interest, we achieve a faster and more efficient reconstruction process without compromising the quality of the final results.

\vspace{1mm}
\noindent\textbf{Dynamic SDF caching.} 
To efficiently find a ray-surface intersection, we search for the first sign flip of the Signed Distance Function (SDF) among a set of $N_c$ equally-spaced points along the ray. However, to reduce the number of MLP queries, we implement a caching mechanism using a voxel grid. When we sample a point belonging to a voxel with a cached SDF value $s$, we compare it with a threshold $\epsilon$. If $s \geq \epsilon$, we consider the point to be far from the surface and safely reuse the cached value. Otherwise, we re-evaluate the SDF network and update the cache accordingly.

To prevent deadlocks, where the cached SDF value remains unjustifiably greater than $\epsilon$, we introduce a random forcing mechanism. With a probability $p$, we deliberately trigger the evaluation and re-caching of the SDF network. After locating the interval where the first SDF sign flip occurs, we further refine the intersection estimation by repeating this process with a set of $N_f$ sub-sampled points.

\begin{figure}[t]
    \centering
    \includegraphics[width=\columnwidth]{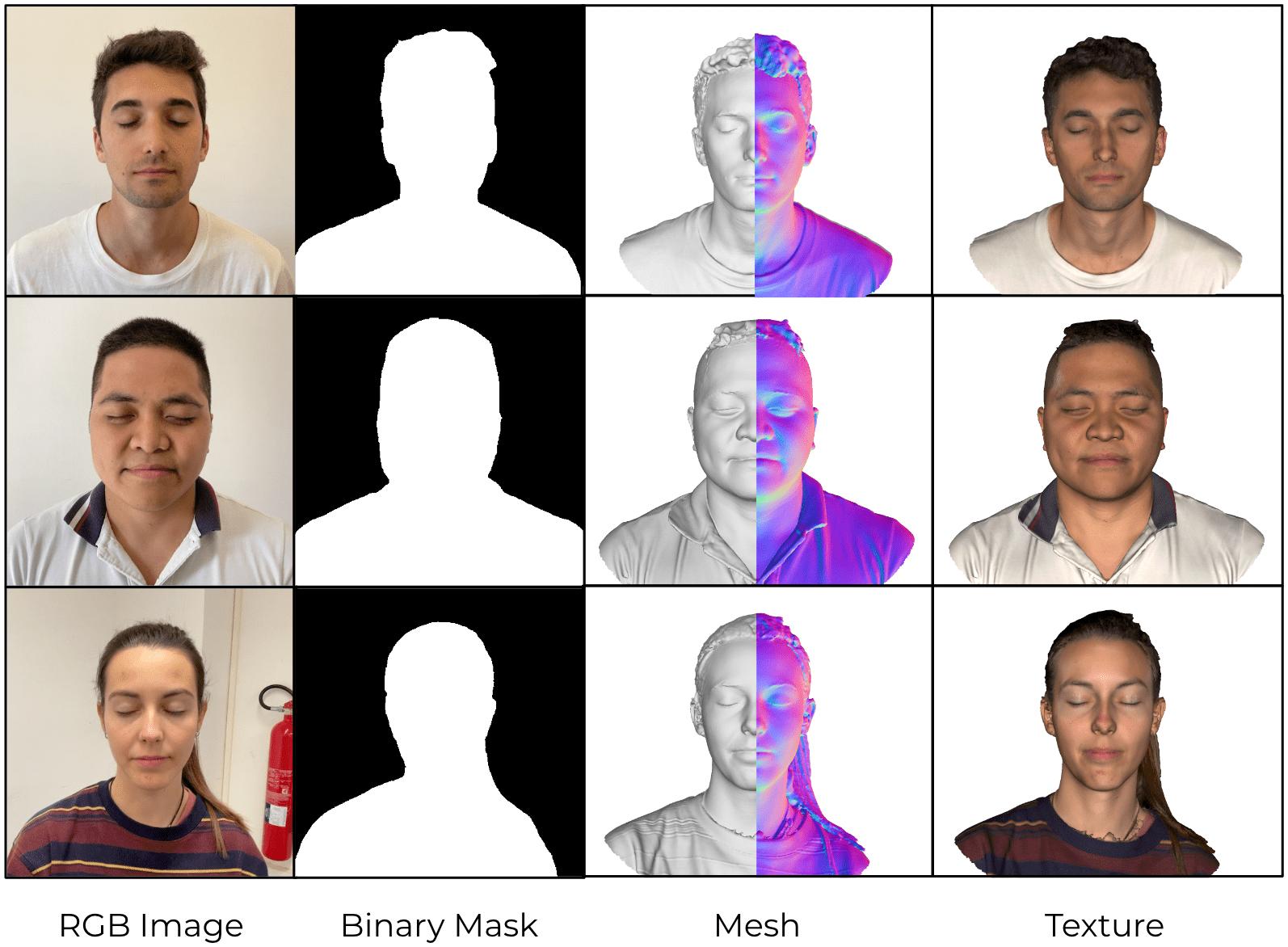}
    \vspace{-6mm}
    \caption{{\bfseries H3DS Dataset.} 
    Three samples from the dataset, each scene composed of 60-100 RGB images, foreground masks, camera parameters, and high-resolution textured 3D meshes capturing the full head, including hair and upper body clothing}.
    \label{fig:h3ds_dataset}
    \vspace{-4mm}
\end{figure}

\section{Implementation details}
\label{sec:implementation}

Equations~\ref{eq:one_shot_def_net}, ~\ref{one_shot_canonical_coord} and~\ref{eq:one_shot_geo_net} are implemented using Multi-Layer Perceptrons (MLPs) with one skip connection from the network's input to the input of a hidden layer, following the approach used in \cite{Park_2019_CVPR}. We incorporate a SoftPlus activation function in all the hidden layers of the network architecture. Additionally, positional encoding (PE) \cite{tancik2020fourier} is applied to some of the inputs of the networks, further enhancing their representation capabilities.

The SA-SM pretraining optimization is iterated for 100 epochs using the Adam optimizer~\cite{kingma2014adam} with standard parameters. We set the learning rate to $10^{-4}$ and apply a learning rate step decay of 0.5 every 15 epochs. To balance the different components of the loss function, we set the loss hyperparameters as follows: $\lambda_1 = 0.1$, $\lambda_2= \lambda_3 = \lambda_5 = 10^{-3}$, and $\lambda_4 = 1$. Additionally, we automatically annotate six 3D facial landmarks for each scene, which are then utilized for the landmark consistency loss.

The weights of the reference SDF network (Eq.~\ref{one_shot_canonical_coord}) are initialized using the geometric initialization method described in~\cite{atzmon2020sal}. As for the deformation and rendering networks, their weights are initialized as multivariate Gaussians with zero mean and variance $10^{-4}$. Furthermore, the latent vectors $\bz_{\rm sdf}$ and $\bz_{\rm r}$ are initialized as zero vectors.

We adopt a progressive masking strategy for the positional encoding (PE) of the input to the reference SDF~\cite{lin2021barf, Park_2021_ICCV, hertz2021sape} to minimize artifacts on the reference shape and improve training stability. This technique involves initially masking the higher frequency bands, effectively acting as a dynamic low-pass filter. By allowing the model to focus on reaching robust coarse solutions first before incorporating high-frequency content, we achieve better convergence behavior.
To implement this approach, we introduce a parameter $\zeta \in [0, L]$ that is proportional to the progress of the training, where $L$ represents the total number of frequencies used in the PE. The Fourier embedding of frequency $k$ is then multiplied by a scalar $w_k(\zeta)$, defined as:
\begin{equation}\label{eq:i3dmm_barf}
    w_k(\zeta) =  \begin{cases} 0 &    \zeta \leq k \\ 
              (1-\cos{(\zeta - k)\pi})/2 &   0 \leq \zeta - k \leq 1 \\
               1 &   \zeta - k \geq 1
             \end{cases}.
\end{equation}

\begin{figure}[t]
    \centering
    \includegraphics[width=\columnwidth]{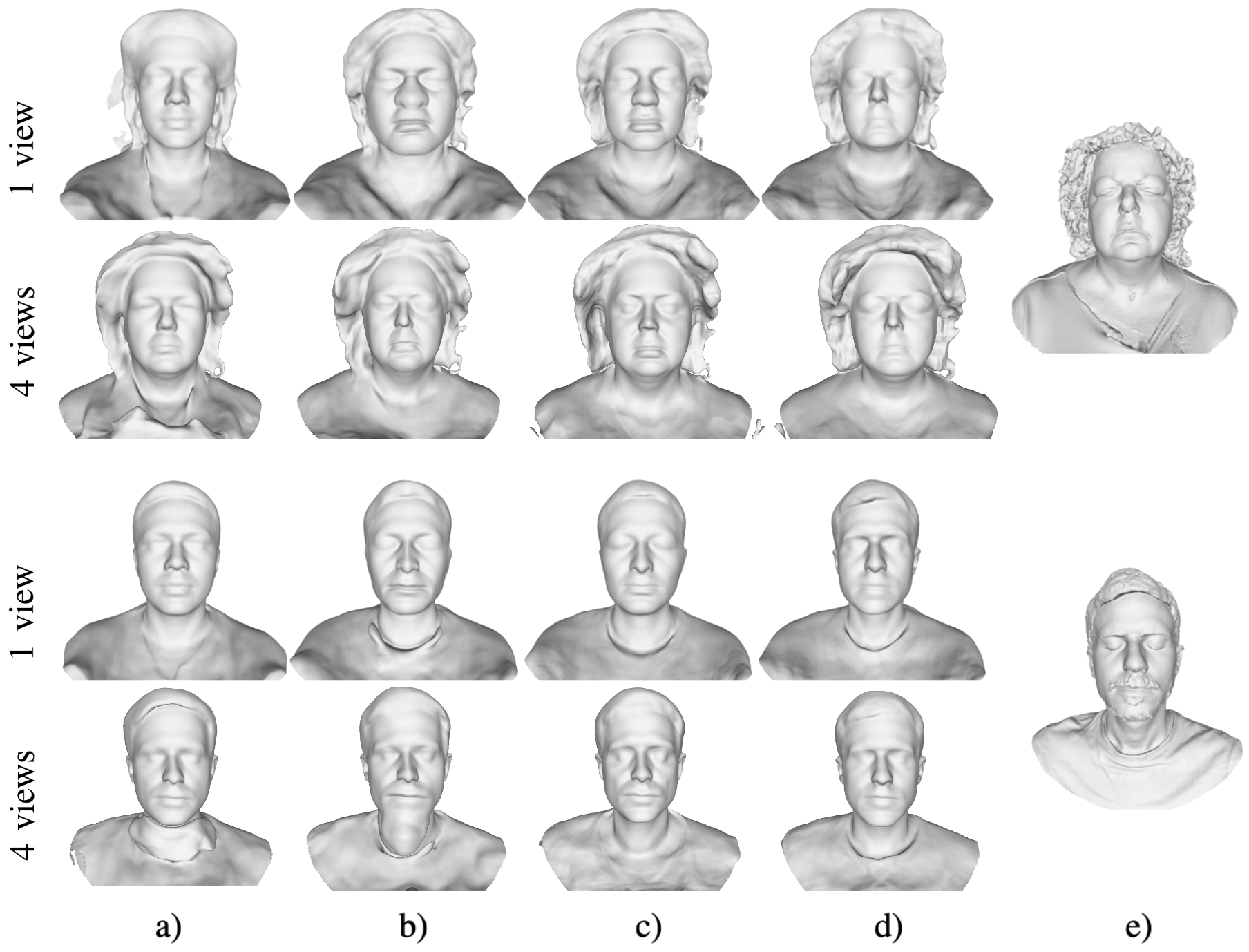}
    \vspace{-6mm}
    \caption{
        {\bfseries Ablation study:} 3D reconstruction  for two subjects from the H3DS dataset. (a) H3D-Net~\cite{ramon2021h3d}; (b) is H3D-Net + progressive masking;  (c) H3D-Net + Deformation Field + Reference SDF; (d) SIRA++ (Ours); (e) Ground Truth. See text for details.
    }
    \label{fig:h3dnet2sira}
    \vspace{-4mm}
\end{figure}

We start with masking all frequencies in the PE and gradually unmask them between epochs 5 and 10 by linearly increasing the parameter $\zeta$ from 0 to $L$.

\vspace{1mm}
\noindent\textbf{Reconstructing geometry from images:} 

At test time, the 3D reconstruction of a scene is performed over 2000 epochs using the Adam optimizer with an initial learning rate of $10^{-4}$. We apply a learning rate step decay of 0.5 at epochs 1000 and 1500 to adaptively adjust the learning rate during the optimization process.

For the mask loss $\mathcal{L}_{\rm Mask}$, we schedule the parameter $\lambda_{8}$ following the approach in~\cite{yariv2020multiview}. We use a two-step scheduling strategy, where the weights of the deformation and rendering networks are unfrozen at epoch 100, enabling more focused fine-tuning during the latter stage of the optimization.

During optimization, we drop 12\% of the background rays every 250 epochs during the initial 1000 epochs. For caching, we implement a voxel grid with a size of $64^3$. We set $\epsilon$ to 0.1 and we randomly sample points along the ray with a probability of $p=0.2$ to ensure exploration and avoid deadlocks. The ray sampling is implemented using $N_c = 75$ and $N_f = 25$ steps.

\section{Experiments}

\subsection{Datasets}

\begin{table}[t!]
\setlength{\tabcolsep}{4pt} % general space between cols (6pt standard)
\centering
\vspace{-2mm}
\caption{{\bfseries Ablation study}  in the one-shot (1 view) and 4 views setup. The face and full-head mean distances are the averages over the 23 subjects in the H3DS dataset in mm. The configurations a, b, c, and d are the same as those described in Figure \ref{fig:h3dnet2sira}}
\resizebox{0.5\textwidth}{!}{
\sisetup{detect-weight=true}
\begin{tabular}{ccccc}
\toprule
& (a) & (b) & (c) & (d)\\
\cmidrule{2-5}
Face mean distance (1 view) & 1.97 & 2.17 & 1.86 & \bfseries 1.46 \\
Full-head mean distance (1 view) & 15.40 & 14.00 & \bfseries 13.40 & 14.16 \\
\cmidrule{1-5}
Face mean distance (4 views) & 1.49 & 1.54 & 1.38 & \bfseries 1.29 \\
Full-head mean distance (4 views) & 11.38 & 10.61 & \bfseries 9.00 & 10.51 \\
\bottomrule
\end{tabular}
}
\vspace{2mm}
\label{table:ablation_h3dnet2sira}
\vspace{-4mm}
\end{table}

\vspace{-1mm}
\noindent\textbf{Prior training.}
To train the geometry prior, we utilize an internal dataset comprising 3D head scans from 10,000 individuals. This dataset is intentionally designed to be well-balanced in terms of gender representation and diverse in terms of age and ethnicity. Before training, the raw data undergoes an automatic processing step to remove internal mesh faces and non-human parts, such as background walls. To ensure consistency and alignment across the dataset, all the scenes are registered by using a non-rigid Iterative Closest Point (ICP) approach to align each head scan with a template 3D model.

\begin{table}[t!]
\begin{center}
\centering
\caption{Camera noise comparison. Evaluation on H3DS dataset with 3 input views.}
\label{tab:camera_error}
\begin{tabular}[t]{ccccc}
\toprule
& Noise & Noise & face & head \\
& & Error ($\sigma$) & (mm) $\downarrow$ & (mm) $\downarrow$\\
\cmidrule{1-5}

H3D-Net \cite{ramon2021h3d} & \multirow{2}{*}{\ding{55}} & \multirow{2}{*}{-}  & 1.44 & 11.6 \\
ours & & & \bfseries 1.30 & \bfseries 10.59 \\

\cmidrule{1-5}

H3D-Net \cite{ramon2021h3d} & \multirow{2}{*}{\ding{51}} & \multirow{2}{*}{0.002}  & 1.51 & 12.35 \\
ours &&& \bfseries 1.30 & \bfseries 10.62 \\
\cmidrule{1-5}
H3D-Net \cite{ramon2021h3d} & \multirow{2}{*}{\ding{51}} & \multirow{2}{*}{0.01}  & 1.60 & 12.36 \\
ours &&& \bfseries 1.37 & \bfseries 10.74 \\
\cmidrule{1-5}
H3D-Net \cite{ramon2021h3d} & \multirow{2}{*}{\ding{51}} & \multirow{2}{*}{0.02}  & 1.71 & 12.96 \\
ours &&& \bfseries 1.48 & \bfseries 10.83 \\
\cmidrule{1-5}
H3D-Net \cite{ramon2021h3d} & \multirow{2}{*}{\ding{51}} & \multirow{2}{*}{0.04} & 1.87 & 13.36 \\
ours &&& \bfseries 1.71 & \bfseries 11.84 \\

\bottomrule
\end{tabular}
\end{center}
\vspace{-4mm}
\end{table}

\vspace{1mm}
\noindent\textbf{H3DS.} There exist several 3D face datasets \cite{zhu2023facescape, wang2022faceverse, Dai2020-cg, RingNet:CVPR:2019, Chen2014-hf, pillai20192nd} that can be used for various tasks, however, large-scale datasets containing high-quality 3D data of full head scans, including hair and shoulders, paired with casual posed RGB images are currently scarce. To address this limitation, we have significantly expanded the H3Ds dataset \cite{ramon2021h3d} by tripling the number of scenes, resulting in a total of 60 subjects. Each subject in the dataset is represented by approximately 100 RGB photos with a resolution of 512x512 pixels, capturing a full 360-degree view around the head. These RGB images are accompanied by foreground masks and camera parameters. Moreover, to enable accurate and reliable ground truth evaluation, the dataset includes high-quality 3D textured scans for each subject. These 3D scans are composed of approximately 150,000 vertices and 400,000 faces, complemented by a texture map with a resolution of 2048x2048 pixels (see fig. \ref{fig:h3ds_dataset}). This dataset can be used either for optimization-based methods or for validation purposes.

The data acquisition process for each scene in the dataset involves several steps. Initially, the camera of an iPad Pro is calibrated using an attached structured light sensor, specifically the Occipital Structure Sensor Pro. This calibration process allows us to obtain paired RGB images and camera parameters, along with a low-resolution mesh scan. Simultaneously, a high-end Artec Eva scanner is utilized to capture high-quality 3D scans. Subsequently, we align the low and high-resolution meshes, along with the paired cameras, by employing six manually annotated 3D landmarks and utilizing the iterative closest point (ICP) refinement technique. Furthermore, for each image within the dataset, we manually annotate a foreground mask, providing additional information for foreground-background separation.

\noindent\textbf{3DFAW. \cite{pillai20192nd}} This dataset provides videos recorded as well as mid-resolution 3D ground truth of the facial region. We select 5 male and 5 female scenes and use them to evaluate only the facial region.

\subsection{Ablation Study} \label{ablation}

For the ablation study, we focus on a subset of 23 scenes, randomly chosen, from the H3DS dataset.

\definecolor{rowblue}{RGB}{220,230,240}
\begin{table*}[t!]
\setlength{\tabcolsep}{4pt} % general space between cols (6pt standard)
\centering
\rowcolors{6}{rowblue}{white}
\vspace{-2mm}
\caption{{\bfseries 3D reconstruction comparison.} Average surface error (in mm) computed over all subjects in   3DFAW and H3DS datasets. We place "-" for not applicable configurations and "$\ast$" for experiments that raised out of memory error.}
\resizebox{0.95\textwidth}{!}{
\sisetup{detect-weight=true}
\begin{tabular}{lcccccccccccccccccc}
\toprule
& \multicolumn{4}{c}{3DFAW} & \multicolumn{12}{c}{H3DS 2.0} \\
\cmidrule(lr{0.5em}){2-5} \cmidrule(lr{0.5em}){6-19}
& \multicolumn{2}{c}{1 view} & \multicolumn{2}{c}{3 view} & \multicolumn{2}{c}{1 views} & \multicolumn{2}{c}{3 views} & \multicolumn{2}{c}{4 views} & \multicolumn{2}{c}{6 views} & \multicolumn{2}{c}{8 views} & \multicolumn{2}{c}{16 views} & \multicolumn{2}{c}{32 views}  \\
\cmidrule{1-19}
& \multicolumn{2}{c}{face} & \multicolumn{2}{c}{face} & face & head & face & head & face & head & face & head & face & head & face & head & face & head \\
MVFNet \cite{wu2019mvf} & \multicolumn{2}{c}{-}& \multicolumn{2}{c}{1.56} & - & - & 1.73 & - & - &- & - & - & - & - & - & - & - & -\\
DFNRMVS \cite{bai2020deep} & \multicolumn{2}{c}{-}& \multicolumn{2}{c}{1.69}  & - & - & 1.83 & - & - &- & - & - & - & - & - & - & - & -\\
DECA \cite{feng2021learning} & \multicolumn{2}{c}{1.71}& \multicolumn{2}{c}{-}  & 1.99 & - & - & - & - &- & - & - & - & - & - & - & - & -\\
MICA \cite{MICA:ECCV2022} & \multicolumn{2}{c}{1.83}& \multicolumn{2}{c}{-}  & 2.08 & - & - & - & - &- & - & - & - & - & - & - & - & -\\
FaceVerse \cite{wang2022faceverse} & \multicolumn{2}{c}{1.88}& \multicolumn{2}{c}{-}  & 2.57 & - & - & - & - &- & - & - & - & - & - & - & - & -\\
FaceScape \cite{zhu2023facescape} & \multicolumn{2}{c}{1.61}& \multicolumn{2}{c}{-}  & 1.78 & - & - & - & - &- & - & - & - & - & - & - & - & -\\
HRN \cite{lei2023hierarchical}  & \multicolumn{2}{c}{1.60}& \multicolumn{2}{c}{-}  & 1.73 & - & - & - & - &- & - & - & - & - & - & - & - & -\\
PIFU \cite{saito2019pifu} & \multicolumn{2}{c}{2.19}& \multicolumn{2}{c}{1.99}  & 1.98 & 12.6 & 1.70 & 11.3 & 1.85 & 11.8 & 2.03 & 10.9 & $\ast$ & $\ast$ & $\ast$ & $\ast$ & $\ast$ & $\ast$\\
JIFF \cite{cao2022jiff} & \multicolumn{2}{c}{1.48}& \multicolumn{2}{c}{1.47}  & 1.85 & 11.5 & 1.80 & 11.7 & 1.79 & 11.2 & 1.79 & 10.9 & $\ast$ & $\ast$ & $\ast$ & $\ast$ & $\ast$ & $\ast$\\
IDR \cite{yariv2020multiview} & \multicolumn{2}{c}{-}& \multicolumn{2}{c}{3.92}  & - & - & 3.51 & 28.8 & 3.33 & 14.3 & 3.12 & 16.4 & 2.97 & 12.6 & 2.88 & 12.4 & 2.04 & 10.6\\
NeuS2 \cite{neus2}  & \multicolumn{2}{c}{-} & \multicolumn{2}{c}{-} & - & - & - & - & 3.96 & 8.29 & 2.18 & 6.60 & 2.70 & 5.79 & 2.11 & 4.55 & 1.85 & 4.14\\
H3D-Net \cite{ramon2021h3d} & \multicolumn{2}{c}{1.70} & \multicolumn{2}{c}{1.37} & - & - & 1.44 & 11.6 & 1.41 & 8.32 & 1.21 & 5.75 & 1.35 & 7.88 & 1.17 & 6.91 & 1.04 & 5.67\\
SIRA++ (Ours) & \multicolumn{2}{c}{\bfseries 1.35} & \multicolumn{2}{c}{\bfseries 1.32} & \bfseries 1.57 & \bfseries 10.79 & \bfseries 1.18 & \bfseries 8.63 & \bfseries 1.23 & \bfseries 5.31 & \bfseries 1.04 & \bfseries 4.81 & \bfseries 1.18 &  \bfseries  4.63 & \bfseries 1.07 & \bfseries 4.26 & \bfseries 1.02 & \bfseries 4.12\\
\bottomrule
\end{tabular}
}
\label{table:quantitative}
\vspace{-2mm}
\end{table*}

\begin{figure*}[t!]
    \centering
    \includegraphics[width=1.\textwidth]{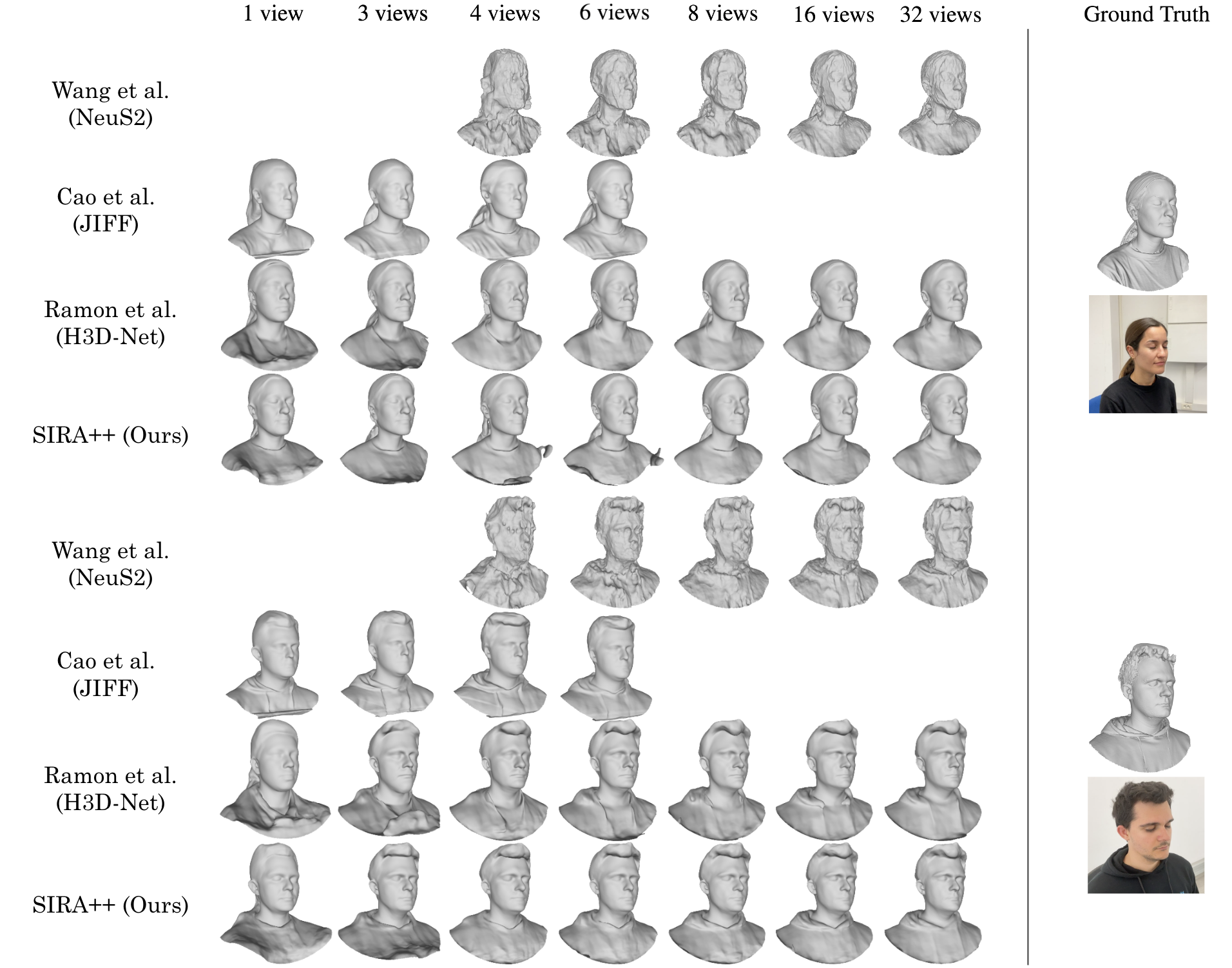}
    \vspace{-6mm}
    \caption{{\bfseries Qualitative results on two subjects of the H3DS dataset}, for NeuS2~\cite{neus2}, JIFF~\cite{cao2022jiff}, H3D-Net~\cite{ramon2021h3d} and SIRA++,  with an  increasing   number of input views.}
    \label{fig:qualitative_idr}
    \vspace{-4mm}
\end{figure*}

\vspace{1mm}
\noindent\textbf{Architecture analysis.} 
\method{} architecture (see sec. \ref{sec:method}) introduces two significant differences compared to H3D-Net~\cite{ramon2021h3d}: it represents the geometry as a deformed reference SDF and incorporates pretraining for rendering human head appearances. We have found both of these strategies to be crucial for achieving high-quality results, especially when the number of views decreases. To thoroughly investigate their impact, we conduct the ablation study on the H3DS dataset for both the one-shot and three-shot scenarios. The qualitative results are presented in Figure~\ref{fig:h3dnet2sira}.

We utilize H3D-Net \cite{ramon2021h3d} as our baseline. As evident from Fig.~\ref{fig:h3dnet2sira}, this architecture underfits the scene when only the latent vector is optimized (a), and in (b) it becomes unstable when its unique geometry decoder is fine-tuned following the progressive masking of Eq.~\ref{eq:i3dmm_barf}. To address these issues, we introduce a significant enhancement by splitting the geometry into a deformation field and a reference SDF (c). This modification leads to more plausible and stable solutions.
Furthermore, \method{} (d) leverages joint modeling of the distribution of 3D shapes and appearances with the \priorgeo{}, enabling better disambiguation of geometric and visual information. Consequently, the 3D models generated by \method{} highly resemble the input images in (e). The quantitative results of this study, over the 23 scenes are reported in Table~\ref{table:ablation_h3dnet2sira}.

\begin{figure*}[t!]
    \centering
    \includegraphics[width=1.\textwidth]{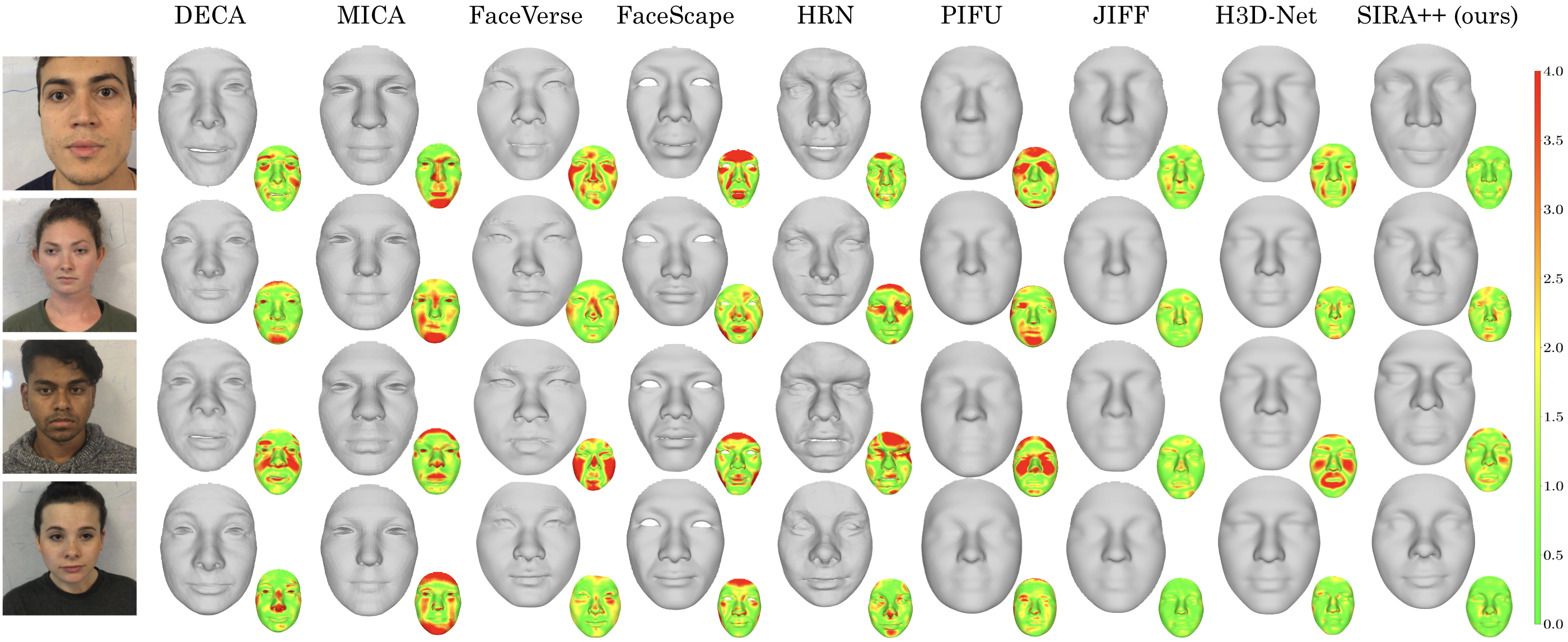}
    \vspace{-4mm}
    \caption{
    \textbf{Qualitative results on the 3DFAW dataset for a single input image.} Each 3D reconstructed face is accompanied by a heatmap, where reddish areas indicate larger errors in mm.}
    
    \label{fig:3dmm_1_view}
    \vspace{-2mm}
\end{figure*}

\begin{figure*}[t!]
    \centering
    \includegraphics[width=1.\textwidth]{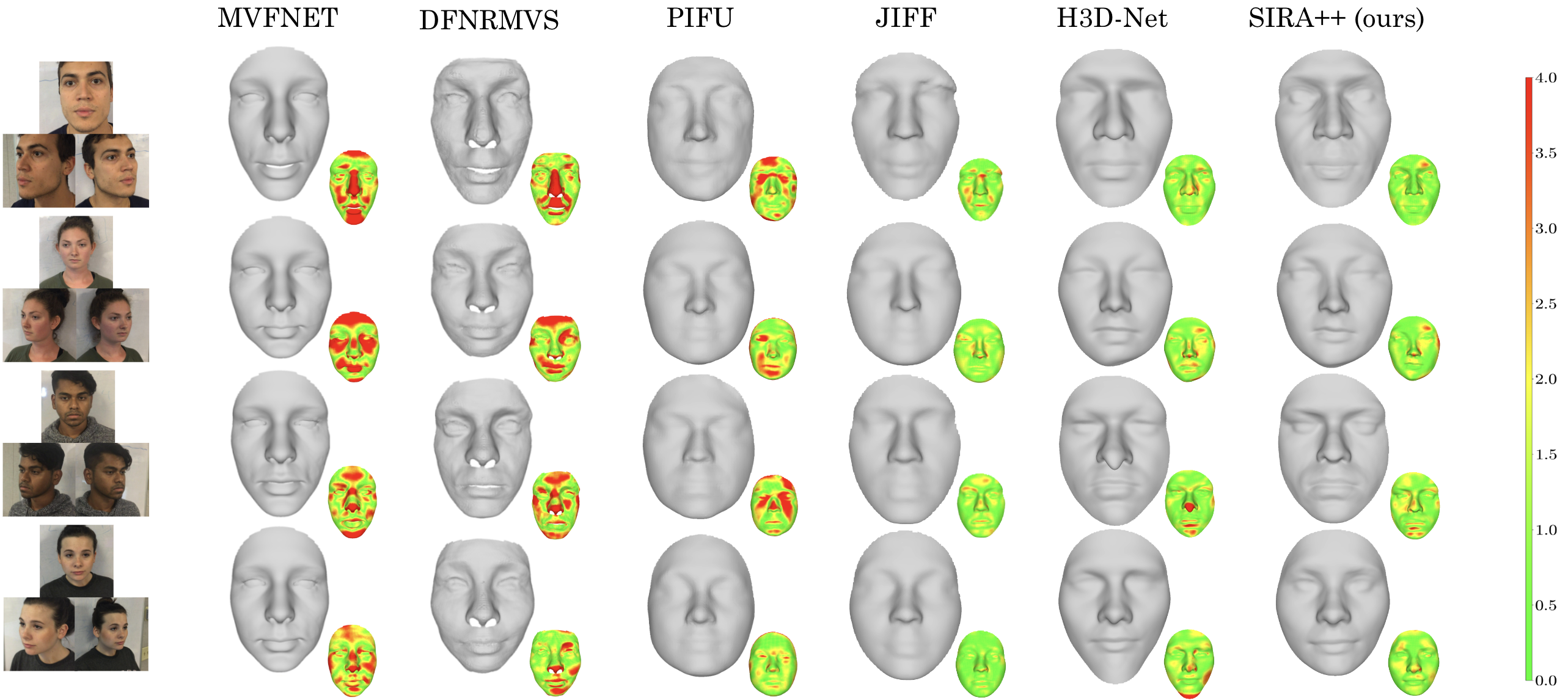}
    \vspace{-4mm}
    \caption{\textbf{Qualitative results on the 3DFAW dataset for three input images.} Each 3D reconstructed face is accompanied by a heatmap, where reddish areas indicate larger errors in mm.}
    \label{fig:3dmm_3_view}
    \vspace{-2mm}
\end{figure*}

\vspace{1mm}
\noindent\textbf{Robustness to camera noise.}
In real-world scenarios, a certain level of inaccuracy in camera poses is inevitable, leading to multi-view inconsistencies. To evaluate the robustness of \method{} in such situations, we conducted an ablation study by introducing varying levels of noise into the camera poses. We applied different levels of Gaussian noise, where the standard deviation $\sigma$ controlled the amount of noise injected into the rotation matrix and position of the camera.  Remarkably, our method demonstrates strong resilience against the injected noise, and consistently betten than~\cite{ramon2021h3d}. The results of this study are presented in Table \ref{tab:camera_error}.

\subsection{Quantitative results}
We conducted a comprehensive comparison of our method with several 3DMM-based reconstruction works, including MVFNet \cite{bai2020deep}, DFNRMVS \cite{wu2019mvf}, DECA \cite{feng2021learning}, MICA \cite{MICA:ECCV2022}, FaceScape \cite{zhu2023facescape}, FaceVerse \cite{wang2022faceverse} and HRN \cite{lei2023hierarchical}. Additionally, we compared our approach to the model-free methods IDR \cite{yariv2020multiview}, NeuS2 \cite{neus2}, PIFU \cite{saito2019pifu}, JIFF \cite{cao2022jiff} and H3D-Net \cite{ramon2021h3d}. For the quantitative evaluation, we used the unidirectional Chamfer distance, measuring the surface error from the ground truth to the predictions. The results of this comparison are summarized in Table \ref{table:quantitative}.

Both model-free methods (H3D-Net and \method{}) outperform the 3DMM-based methods for all the evaluated view configurations. Notably, the enhancement due to the prior in \method{} becomes more significant as the number of views decreases. However, the prior does not hinder the model from also becoming more accurate when more views are available, which is a limitation in 3DMM-based approaches. In the one-shot regime, \method{} stands out over 3DMM-based approaches and H3D-Net, all of which yield similar results. Both single feed-forward PIFU and JIFF methods generate convincing and robust results, although reconstructions are smooth and they are not able to capture high-frequency details, especially in the face region. \method{} consistently outperforms IDR, NeuS2 and H3D-Net for all configurations, demonstrating its improved ability to generalize under data scarcity. When more views become available and the task is more constrained, \method{} and H3D-Net converge towards comparable performance, as the prior knowledge becomes less critical for obtaining plausible results. These findings are further supported by the qualitative results obtained.

\begin{figure*}[t!]
    \centering
    \includegraphics[width=1.\textwidth]{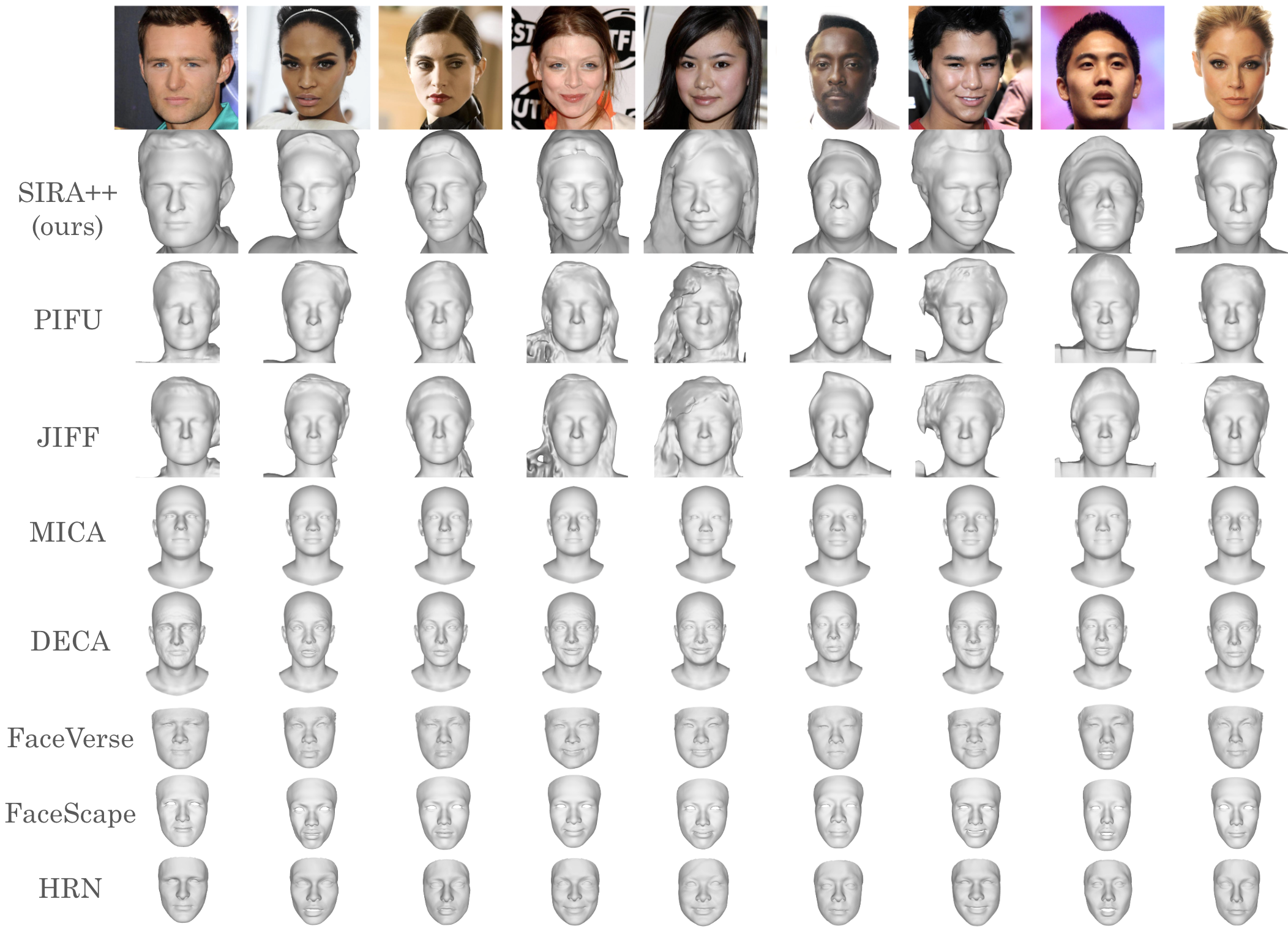}
    \vspace{-6mm}
    \caption{\textbf{Qualitative results on the CelebA-HQ dataset for a single input image.}}
    \label{fig:sira_celeb}
    \vspace{-4mm}
\end{figure*}

\vspace{-4mm}
\subsection{Qualitative results}

\begin{table}[t!]
\begin{center}
\centering
\caption{\textbf{User study: We collect 375 responses from 25 participants to measure visual fidelity of the reconstructions.}} %\francesc{explica la metrica}}
\vspace{-4mm}
\label{tab:user}
\begin{tabular}[t]{ccccc}
\toprule
& NeuS2 & JIFF & H3D-Net & SIRA++ (ours) \\
\cmidrule{1-5}
VF$\uparrow$ & 1.0 & 2.21  & 3.29 & \bfseries3.5 \\

%\cmidrule{1-5}

\bottomrule
\end{tabular}
\end{center}
\vspace{-5mm}
\end{table}

Figure~\ref{fig:qualitative_idr} illustrates the qualitative results of our approach, \method{}, and state-of-the-art approaches NeuS2 \cite{neus2}, JIFF \cite{cao2022jiff} and H3D-Net~\cite{ramon2021h3d}, for two subjects from the H3DS dataset under  an increasing number of input images. Notably, our method, \method{}, demonstrates superior performance in surface reconstruction, yielding surfaces with reduced errors and a more realistic appearance, particularly within the facial region. Even with a smaller number of input views, our approach excels at obtaining accurate and visually appealing results. To quantitatively asess these results, we perform a user study, with 25 human participants, to evaluate visual fidelity (see Table \ref{tab:user}). We present a photo of a subject and renders of reconstruction for each method. We ask the participants to rank them based on visual fidelity (how well the reconstruction captures the details of the person shown on the image). We assign a numeric value between 1 and 4 for each response based on the order. Results show that SIRA++ reconstructions are consistently better perceived as digital representation of full-heads.

Fig.~\ref{fig:3dmm_1_view} showcases results for single input images obtained from the 3DFAW dataset \cite{pillai20192nd}. Our method, SIRA++, is compared against the one-shot methods DECA, MICA, FaceVerse, FaceScape, HRN, PIFU, JIFF, and H3D-Net. Next to each 3D reconstruction, we display a heatmap representing the reconstruction error. Note that SIRA++ outperforms the 3DMM-based methods, DECA, MICA, FaceVerse, FaceScape and HRN, and model-free methods, PIFU, JIFF and H3D-Net significantly. Specifically, it excels in critical regions like the nose and mouth, which are pivotal in defining the unique anatomical features of each individual.

Similarly, Fig.~\ref{fig:3dmm_3_view} provides an analysis for the case of three input images. Here, we compare SIRA++ against MVF-Net~\cite{wu2019mvf}, DFNRMVS~\cite{bai2020deep}, PIFU, JIFF and H3D-Net. Again, the methods based on coordinate-based neural representation,   H3D-Net, and especially our SIRA++, outperform those relying on 3D Morphable Models (MVF-Net and DFRMVS) and single feed-forward models (PIFU and JIFF).

This is further highlighted in our last experiment, summarized in Fig.~\ref{fig:sira_celeb}, where we specifically focus on the most challenging scenario of utilizing just one single and in-the-wild input image, randomly taken from the celebA-HQ dataset~\cite{karras2017progressive}.
In this case SIRA++ also consistently outperforms single feed-forward methods, PIFU and JIFF as well as the 3DMM methods, DECA, MICA FaceVerse, FaceScape and HRN. DECA and MICA struggle to capture fine anatomical details, often leading to similar-looking faces across different scenes. On the other hand, FaceVerse and FaceScape produce biased outputs toward Asian characteristics, as the training data is composed of Asian subjects. Additionally, these approaches are limited to reconstructing only the facial region and fail to recover the hair and shoulders, which significantly impact perception. In contrast, our method excels at capturing not only the facial features but also includes the hair, upper body clothing, and other high-frequency anatomical details, especially in the cheeks and mouth regions. This ability results in 3D shapes that retain the identity of the person, showcasing the unique characteristics of the individual.
\vspace{-2mm}
\section{Limitations and Future Work}

In our experiments, we show that SIRA++ demonstrates significant advancements in quick personalization of pretrained geometry and appearance priors from a few headshot images. However, we believe that there are still several limitations and opportunities for future work. Firstly, our rendering networks are based on a DeepSDF decoder, and future work could focus on combining them with Gaussian Splats for fast, high-quality rendering. Although our approach achieves state-of-the-art reconstruction accuracy within 200 seconds, this may still be prohibitive for real-time applications or scenarios requiring fast processing of multiple scenes. Future work could focus on optimizing the efficiency of the algorithm. Another limitation is the sensitivity of our method to the quality of the input images. Our approach is sensitive to the resolution, angles, occlusion, and information present in the input images. Variations in these factors can significantly affect the reconstruction quality.
\vspace{-2mm}
\section{Conclusions}

In this paper, we have introduced SIRA++, a method for high-fidelity full 3D head reconstruction in few-shot and in-the-wild scenarios. To address the inherent ambiguity of the problem, we proposed a novel statistical model based on neural fields, which encoded shape and appearance into low-dimensional latent spaces. The thorough evaluation demonstrated that our approach achieved state-of-the-art results in full head geometry reconstruction. Moreover, through a detailed ablation study, we showcased the robustness of our method to camera pose misalignment. We also presented a set of improvements that led to an impressive 80\% reduction in computation time compared to previous approaches, H3D-Net~\cite{ramon2021h3d} and SIRA~\cite{Caselles_2023_SIRA}.

% PER AFEGIR QUAN ACCEPTIN EL PAPER
%\section*{Acknowledgments}
%This work has been partially funded by the Spanish government with the projects MoHuCo PID2020-120049RB-I00 and DeeLight PID2020-117142GB-I00 funded by MCIN/ AEI /10.13039/501100011033, and the industrial doctorate 2021 DI 17 funded by the Government of Catalonia.

\bibliography{egbib}
\bibliographystyle{IEEEtran}

\vspace{-7mm} 
\begin{IEEEbiography}
[{\includegraphics[width=1in,height=1.25in,clip,keepaspectratio]{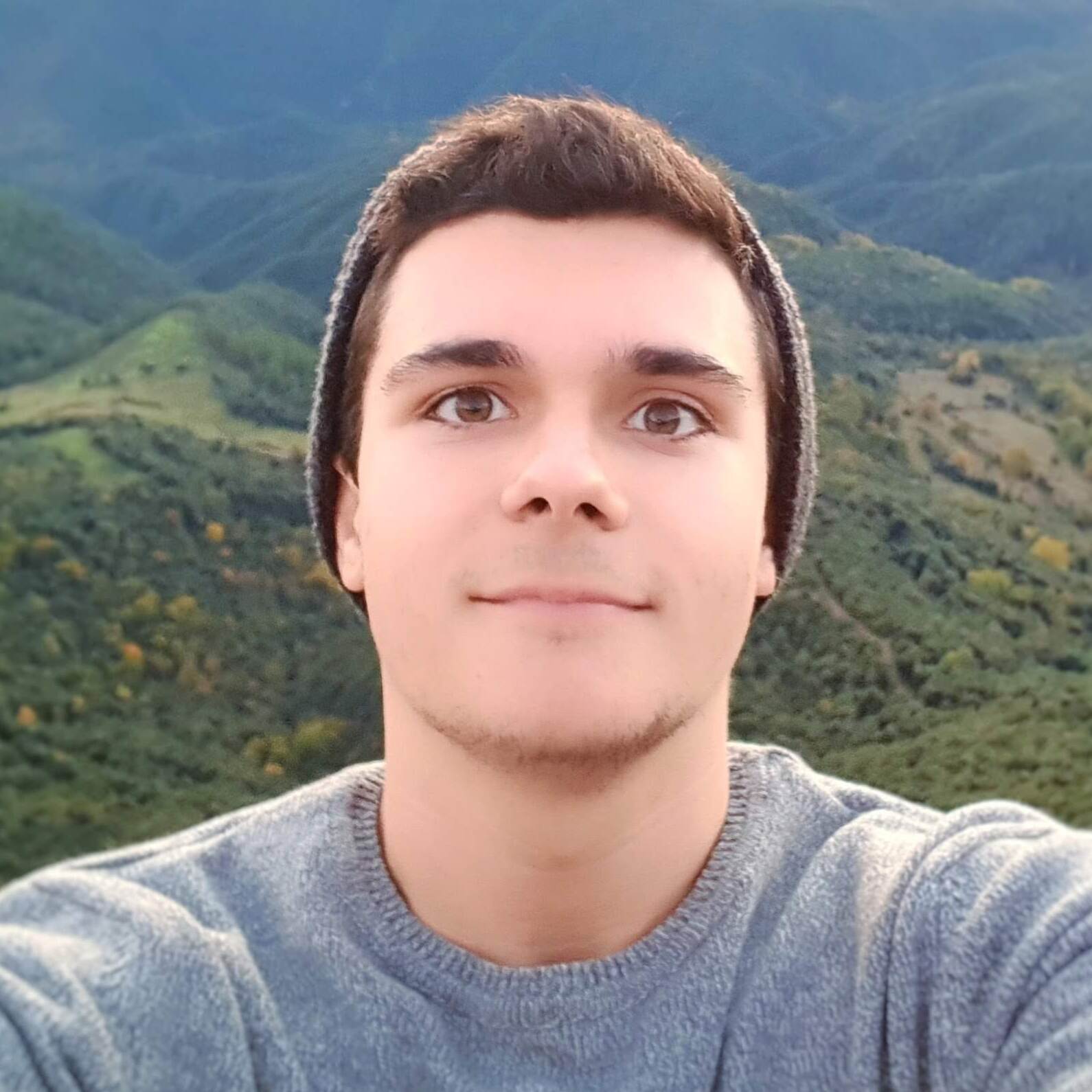}}]{Pol Caselles Rico} is a third-year industrial PhD candidate at Universitat Politècnica de Catalunya (UPC). He is currently working as an applied scientist at Crisalix Labs in Barcelona. His primary research interests are in Computer Vision and Machine Learning, with a specific emphasis on 3D shape reconstruction from 2D images, using implicit functions and model-free approaches.
\end{IEEEbiography}

\vspace{-15mm}
\begin{IEEEbiography}
[{\includegraphics[width=1in,height=1.25in,clip,keepaspectratio]{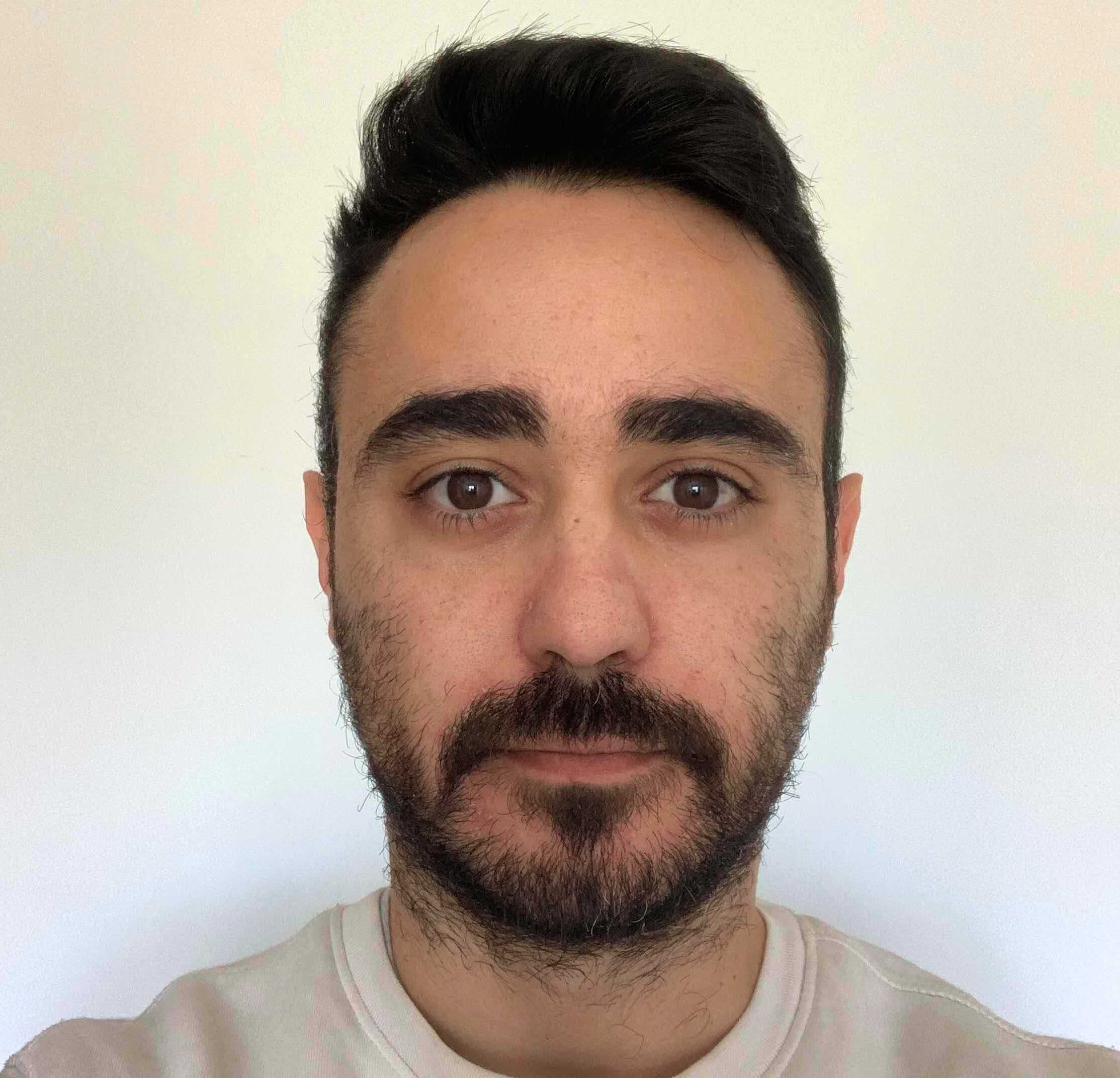}}]{Eduard Ramon} received his PhD in computer science from the Universitat Politècnica de Catalunya in 2022. During his PhD and previously, he worked at Crisalix as a computer vision scientist and (co)authored several publications on topics related to statistical models, and 3D reconstruction of human bodies and faces. Currently, he is working as an applied scientist at Amazon.
\end{IEEEbiography}

\vspace{-15mm}
\begin{IEEEbiography}
[{\includegraphics[width=1in,height=1.25in,clip,keepaspectratio]{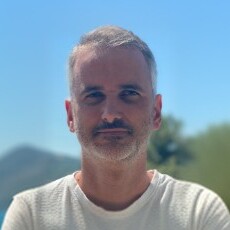}}]{Jaime Garcia Giraldez} is the CEO and founder of Crisalix, Switzerland. He received his PhD in Biomedical Engineering from the Medical University of Bern. He has co-authored more than 10 publications in refereed journals and conferences. His research interests are in computer graphics, 3D and Augmented Reality, Computer Assisted Surgery (CAS), computer vision and AI.
\end{IEEEbiography}

\vspace{-15mm}
\begin{IEEEbiography}
[{\includegraphics[width=1in,height=1.25in,clip,keepaspectratio]{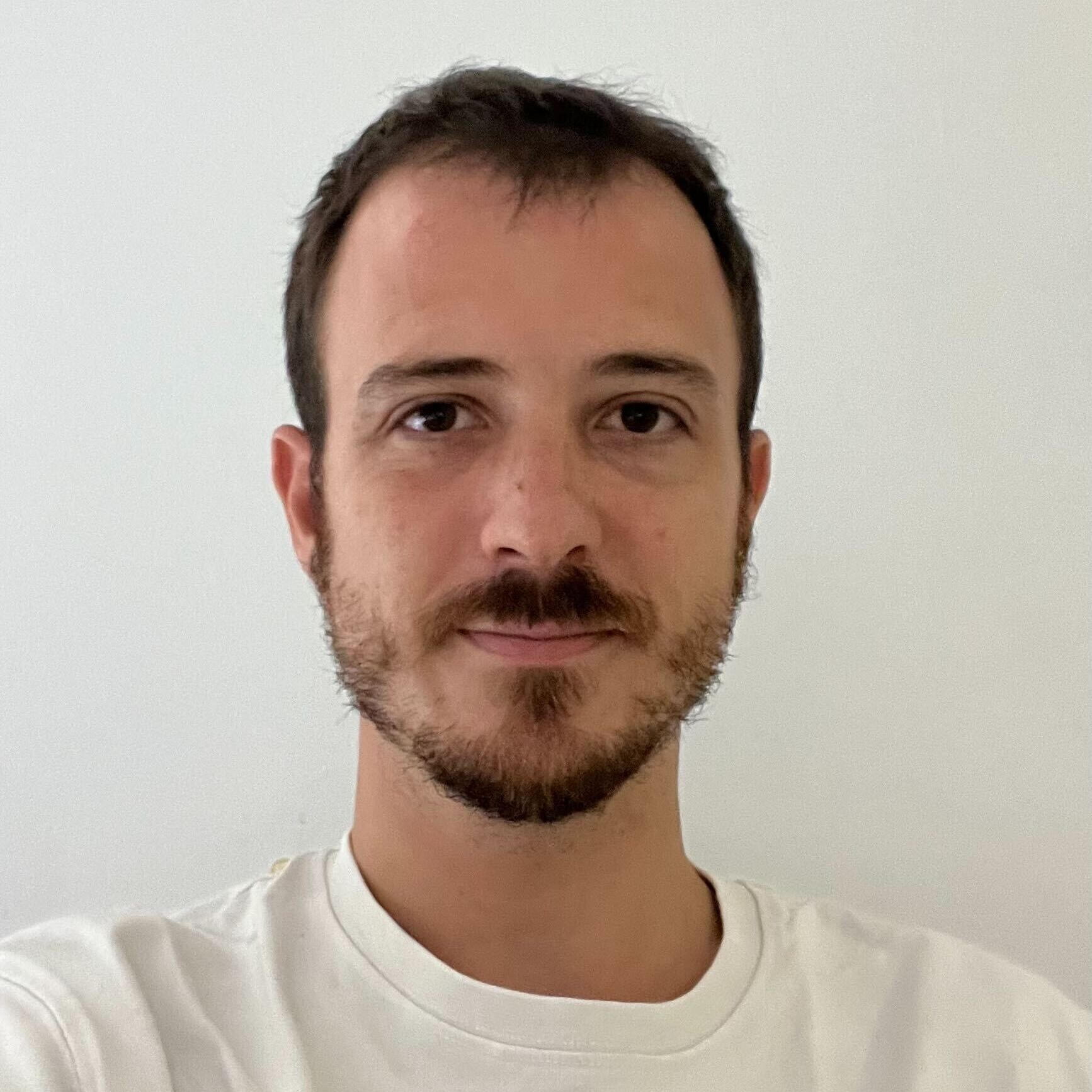}}]{Gil Triginer} received his PhD in physics from the University of Oxford in 2019. Subsequently, he joined Crisalix Labs as a research scientist, with a focus on 3D reconstruction of human faces and bodies using deep learning techniques. He has coauthored publications in leading computer vision conferences including ICCV, WACV, and ACCV workshops.
\end{IEEEbiography}

\vspace{-15mm}
\begin{IEEEbiography}
[{\includegraphics[width=1in,height=1.25in,clip,keepaspectratio]{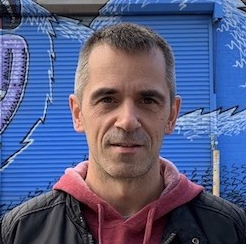}}]{Francesc Moreno-Noguer}  is a Principal Applied Scientist at Amazon Science, specializing in Computer Vision and Machine Learning. His research focuses on human shape and motion estimation, 3D reconstruction of both rigid and nonrigid objects, and camera calibration. He received the Polytechnic University of Catalonia’s Doctoral Dissertation Extraordinary Award, multiple best paper awards (e.g. ECCV 2018 Honorable mention, ICCV 2017 workshop in Fashion, Intl. Conf. on Machine Vision applications 2016), outstanding reviewer awards at ECCV 2012 and CVPR 2014, and Google and Amazon Faculty Research Awards in 2017 and 2019, respectively. He has (co)authored over 200 publications in refereed journals and conferences (including 13 IEEE Transactions on PAMI, 5 Intl. Journal of Computer Vision, 30 CVPR, 13 ECCV and 10 ICCV).
\end{IEEEbiography}

\vfill

\end{document}